\documentclass[journal]{IEEEtran}

\usepackage{newtxtext,newtxmath}

\usepackage{moreverb,url}
\usepackage{amsmath}
\usepackage{bm}

\usepackage[acronym]{glossaries}
\usepackage{cases}
\usepackage{amssymb}
\usepackage{graphicx}
\usepackage{subfigure}
\usepackage{multirow}
\usepackage{caption2}
\usepackage{cite}
\usepackage[noend]{algpseudocode}
\usepackage{algorithmicx,algorithm}
\usepackage{makecell}
\usepackage{booktabs}
\usepackage{enumitem}

\begin{document}

\title{A bilevel optimal motion planning (BOMP) model with application to autonomous parking}

\author{Shenglei Shi, Youlun Xiong, Jiankui Chen and Caihua Xiong}

\author{Shenglei Shi, Youlun Xiong, Jiankui Chen and Caihua Xiong,~\IEEEmembership{Member,~IEEE}\\
        \thanks{
         This work was supported by the National Science Foundation of China under Grant 51327801.\newline \indent
         S.~Shi, Y.~Xiong, J.~Chen and C.~Xiong are with the State Key Laboratory of Digital Manufacturing Equipment and Technology, School of Mechanical Science and Engineering, Huazhong University of Science and Technology, 1037 Luoyu Road, Wuhan, 430074, P.R. China (e-mail: shshlei@hust.edu.cn; famt@hust.edu.cn; Chenjk@hust.edu.cn; chxiong@hust.edu.cn).\newline \indent 
         Color versions of one or more of the figures in this paper are available online at http://ieeexplore.ieee.org.\newline \indent
         Digital Object Identifier \dots
        }
}

\markboth{IEEE TRANSACTIONS ON INTELLIGENT TRANSPORTATION SYSTEMS}%
{Bilevel optimal motion planning and control}

\maketitle

\begin{abstract}
    In this paper, we present a bilevel optimal motion planning (BOMP) model for autonomous parking. The BOMP model treats motion planning as an optimal control problem, in which the upper level is designed for vehicle nonlinear dynamics, and the lower level is for geometry collision-free constraints. The significant feature of the BOMP model is that the lower level is a linear programming problem that serves as a constraint for the upper-level problem. That is, an optimal control problem contains an embedded optimization problem as constraints. Traditional optimal control methods cannot solve the BOMP problem directly. Therefore, the modified approximate Karush-Kuhn-Tucker theory is applied to generate a general nonlinear optimal control problem. Then the pseudospectral optimal control method solves the converted problem. Particularly, the lower level is the $J_2$-function that acts as a distance function between convex polyhedron objects. Polyhedrons can approximate vehicles in higher precision than spheres or ellipsoids. Besides, the modified $J_2$-function (MJ) and the active-points based modified $J_2$-function (APMJ) are proposed to reduce the variables number and time complexity. As a result, an iteirative two-stage BOMP algorithm for autonomous parking concerning dynamical feasibility and collision-free property is proposed. The MJ function is used in the initial stage to find an initial collision-free approximate optimal trajectory and the active points, then the APMJ function in the final stage finds out the optimal trajectory. Simulation results and experiment on Turtlebot3 validate the BOMP model, and demonstrate that the computation speed increases almost two orders of magnitude compared with the area criterion based collision avoidance method.
\end{abstract}

\section{Introduction}
\label{sec:introduction}
Real-time collision-free motion planning and control for autonomous vehicles have received a considerable amount of attentions, and share many research methods with robotics literature \cite{lavalle2006planning,gonzalez2016review,vorobieva2015automatic,du2015autonomous,muller2007continuous,xu2018model,liu2017parking,tazaki2017parking,cowlagi2012hierarchical,khatib1986real,oetiker2009navigation,karaman2011sampling,chi2018risk,janson2018deterministic,7452671,lim2018hierarchical,zucker2013chomp,schulman2014motion,park2015homotopy,pan2012collision,robinson2018efficient,li2015unified,li2016time,nilsson2016lane,rasekhipour2016potential,ye2018variable}. 
Typically, autonomous parking is a critical maneuver especially in big narrow cities. Separated methods \cite{vorobieva2015automatic,du2015autonomous,muller2007continuous,xu2018model} are common approaches for  parking problems, that vehicle path planning and path tracking are handled separately. Direct learning \cite{liu2017parking} is a novel and simple approach that learns the mapping relation between control inputs and parking trajectories. However, for complex high-precision parking problems, combined approaches \cite{li2015unified,li2016time} are more effective, that vehicle motion planning and control are treated as a unified optimal control problem. In this paper, we also treat the parking problem as a combined optimal control problem. 

Motion planning approaches can be categorized into graph searching methods \cite{tazaki2017parking,cowlagi2012hierarchical}, artificial potential field methods \cite{khatib1986real,oetiker2009navigation}, sampling-based methods \cite{karaman2011sampling,chi2018risk,janson2018deterministic,7452671,lim2018hierarchical} and nonlinear optimization methods \cite{zucker2013chomp,schulman2014motion,park2015homotopy,pan2012collision,robinson2018efficient,li2015unified,li2016time,nilsson2016lane,rasekhipour2016potential,ye2018variable}. In graph searching methods, systematic discretization of the environment or robot state space is applied to construct a graph first, then highly informative heuristics are used to guide the search to generate a connected feasible path or trajectory. Their practical applications are limited to low dimensional space since the discretization suffers from a large number of discrete cells. Artificial potential field method has made significant achievements in static or dynamic environments. Moreover, the application in automatic parking demonstrates its capability for the nonholonomic system. However, the artificial potential field method is likely to get stuck in a locally optimal path and cannot handle the multi-maneuver parking problem because of local planning characteristic. Sampling-based method randomly generates samples in robot configuration space or robot state space and attempts to connect the new samples to the generated graph or tree using local planning methods. Since sampling-based methods do not represent the obstacles free configuration space explicitly, they are computation effective and suitable for high-dimensional space. However, the smoothness issue and solution quality concerning some constraints and objectives need to be considered seriously \cite{pan2012collision}. On account of the solution quality problem, nonlinear optimization methods can treat robot dynamics correctly, and some concerned objectives and constraints can be constructed to satisfy the task specification. Nevertheless, due to the presence of obstacles, the feasible set is discontinuous, and the nonlinear optimization problem is nonconvex, which increase the difficulty of the problem and the computation complexity.

Optimal control \cite{betts2010practical,bryson2018applied,ross2012review} is a remarkable method to generate high-quality trajectories for robots and has achieved great success in practical applications. It considers robot dynamics and other trajectory constraints in a compact and unified form, and it can deal with any predefined optimization objectives. The application of the optimal control method in robot motion planning involves two essential constraints, robot dynamics constraint, and geometry collision-free constraint. However, the applications are limited to elementary geometries like circles in \cite{robinson2018efficient} and rectangles in \cite{li2015unified}. In this work, we make the optimal control method applicable to complex scenarios with rectangles assumption and solve this problem at high speed. 

\subsection{Problem Analysis}
\label{sec:problemanalysis}
The challenge of making optimal control method applicable in complex motion planning problems arises from how to represent the geometry collision-free constraints effectively. The point-point distance of circles in \cite{robinson2018efficient} and area criterion of rectangles in \cite{li2015unified} are both straightforward collision avoidance methods. However, circle bounding volume approximation is too conservative to realize motion in complex and high-precision scenarios. And area criterion of rectangles is nonlinear and redundant, such that the optimal control problem is difficult to solve. In addition, the collision-free constraints for high-precision motion planning problems in 3D environment are still difficult to be built. We will show in this paper the $J_2$-function \cite{you1987generalized,xiong1989general} settles these challenges, and achieves fast high-precision motion planning for autonomous parking. The $J_2$-function is linear programming for collision checking between convex polyhedrons in any dimensional space, and it behaves as a distance function that $J_2>0$ indicates collision free. Treating $J_2\ge\delta>0$ ($\delta$ is a safety distance, and $J_2$ is a function of robot trajectory) as a constraint for the overall optimal control problem, a special $J_2$-function based bilevel optimal motion planning (BOMP) model is obtained.

The BOMP model contains an optimization problem within the constraints of the upper optimal control problem. Moreover, the lower optimization problem depends on the continuous robot state trajectory such that it is also infinite dimensional. To our knowledge, this particular model has not appeared in the literature, but the complexity of this model can be seen indirectly from some other problems. In \cite{benita2016bilevel}, a relative simple bilevel optimal control problem is considered. The upper level is an optimal control problem concerning ordinary differential equations, control constraints, initial and final state constraints, and the lower level problem depends only on the final state of the physical system and is finite dimensional. In this case, nonsmooth analysis, optimization in Banach spaces and bilevel optimization \cite{sinha2017evolutionary,sinha2018review} are used to derive the necessary linearized Pontryagin-type optimality conditions. By nonconvex, non-differentiable or possibly disconnected, the hierarchical bilevel optimization problem is intrinsically difficult. Even for the most straightforward linear-linear bilevel optimization problem, it was proven to be strong NP-hard \cite{hansen1992new} and that merely evaluating the optimality is also NP-hard \cite{vicente1994descent}. 
   
Two significant challenges are classified from the BOMP model, the bilevel problem, and the optimal control problem. Optimal control has experienced considerable development in mathematics and engineering. In particular, pseudospectral optimal control (PSOC) method \cite{ross2012review} satisfies the differential equations globally and treats integral by an implicit Runge-Kutta method, such that this method achieves high order accuracy, high order stability, and exponential convergence speed. Therefore, the PSOC method is selected. For the bilevel optimization problem, the Karush-Kuhn-Tucker (KKT) reformulation of the lower-level optimization problem is always used to construct a traditional single level optimization problem \cite{albrecht2012bilevel}. The KKT reformulation can not be applied to the BOMP model seemingly since the BOMP problem is an optimal control problem rather than an optimization problem. However, the PSOC algorithm discretizes the whole variables and approximates them as polynomials, such that it actually solves a large scale sparse nonlinear optimization problem (NLP). In this case, the KKT reformulation can be directly used to the discretized NLP. Furthermore, differential equations do not occur in the lower level of the BOMP model, the continuous KKT reformulation (the Lagrange multipliers are also trajectory functions of time) concept can be utilized unambiguously. 
 
However, due to the nonconvexities in the KKT conditions, even the upper-level problem is also convex, the converted problem is still hard to solve. Besides, the complementary constraint is intrinsically combinatorial and makes the extent of violation of KKT conditions in a small neighborhood of the KKT point nonsmooth \cite{dutta2013approximate}. Hence, the convergence peoperty in the neighborhood of the optimal value is poor, and it cannot provide efficient information to determine whether current value is close enough to the optimal value. Dutta \cite{dutta2013approximate} proposed a modified approximate KKT (MAKKT) theory that the KKT conditions are relaxed and the violation in the neighborhood is smooth enough. Since the KKT conditions are relaxed, an iteirative strategy is used to decrease the relaxation factor to approach the optimal value. Therefore, the MAKKT theory and iteirative convergence strategy are selected in this paper.
\subsection{Contributions and Organization}
In this paper, we present a general BOMP model in which the upper level is an optimal control problem designed for robot nonlinear dynamics, while the lower level is the $J_2$-function linear programming for geometry collision-free constraint (in Section \ref{sec:problemdefinition}). The MAKKT theory is used to simplify the model to a traditional optimal control problem, then the PSOC method is used to solve the converted problem (in Section \ref{sec:bomp}). Besides, a modified $J_2$-function (MJ) and an active points based modified $J_2$-function (APMJ) are derived in Section \ref{sec:Jfunction} to reduce time complexity. An iteirative convergence strategy combines the MJ function and APMJ function naturally, and a two-stage BOMP algorithm is proposed. The highlights of this paper are:
\begin{enumerate}[leftmargin=4mm,topsep=2pt,itemsep=0pt,parsep=1pt]
    \item The BOMP model combines both advantages of the flexibility of optimal control and the simplicity of linear programming, which makes optimal control a fast and high-precision method for complex motion planning problems.
    \item For the two-stage algorithm, the MJ function is used in the initial stage to generate an initial collision-free approximate optimal trajectory and to pick out the active points, while the APMJ function is used in the final stage to find out the exact optimal trajectory. 
    \item The BOMP algorithm benefits by the iterative convergence strategy and the simplicities of the MJ function and the APMJ function. Simulations in autonomous parking demonstrate that the computation speed increases almost two orders of magnitude compared with the area criterion based method (in Section \ref{sec:applications}). 
\end{enumerate}

Finally, concluding remarks are given in Section \ref{sec:conclusion} and the issues to be researched in this field are also pointed out.

\section{Bilevel Optimal Motion Planning}
\label{sec:problemdefinition}
Optimization problem finds an optimal point, while optimal control problem finds an optimal trajectory. So it is a natural question: how to use the optimal control method to solve vehicle trajectory generation and optimization problem in complex and high-precision scenarios? As stated in Section \ref{sec:introduction}, the geometry collision-free constraint is the crucial matter. In this section, we will present the BOMP model with an embedded standard linear programming collision avoidance constraint, and the exact $J_2$-function linear programming which acts as a distance function will be presented in section \ref{sec:Jfunction}. 

\begin{figure}[t]
  \centering
  \vspace{-1.5ex}
  \includegraphics[width=0.4\textwidth]{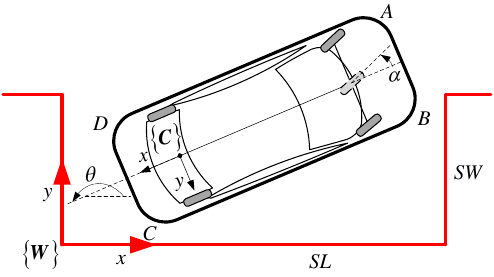}
  \caption{The coordinate systems attached to the vehicle and the parking spot.}
  \label{fig:kinematics}
  \vspace{-1.0ex}
\end{figure}

Vehicle is a typical nonholonomic constrained system, and its motion planning is very complicated. Without loss of generality, the front steering wheels vehicle is considered and the following kinematics model is used:
\begin{equation}
  \left\{
    \begin{aligned}
        & \dot{x}= \upsilon\cos\theta,\quad \ \ \ \dot{y}= \upsilon\sin\theta  \\
        & \dot{\theta} = \upsilon\tan\alpha/l,\, \quad \dot{\alpha} = \omega \\
    \end{aligned}
  \right.
  \label{eq:kinematics}
\end{equation}
where $\bm{q}=(x,\,y,\,\theta)\in \Re^2 \times S$ is the configuration of the vehicle coordinate $\{\bm{C}\}$ with respect to the world coordinate $\{\bm{W}\}$ which origins at one corner point of the parking spot, see Fig. \ref{fig:kinematics}. The vehicle coordinate $\{\bm{C}\}$ origins at the mid-point of the rear wheel axis, $(x,\,y)$ describes the vehicle position, and $\theta$ denotes the orientation. $\upsilon$ denotes the linear velocity of point $(x,\,y)$, $\alpha$ denotes the steering angle of the front wheel and $\omega$ denotes the steering velocity.

Consider vehicle state variable $\bm{x}$ as $\bm{x}=(x,\, y,\, \theta,\, \alpha)$ and the control input $\bm{u}=(\upsilon,\, \omega)$, then vehicle kinematics model (\ref{fig:kinematics}) can be abstracted as $\bm{\dot{x}}=\bm{f}(\bm{x},\,\bm{u})$. The optimal control problem tries to solve an optimal smooth trajectory corresponding to this differential equation constraints and with respect to a specified performance index:
\begin{subequations}
  \begin{align}
      \min \quad &    \beta g(\bm{x}_i,\,\bm{x}_f,\,t_f) + (1-\beta)\int_0^{t_f} L(\bm{x},\,\bm{u}) \mathrm{d}t \nonumber \tag{2} \\ 
      \label{eq:optimalcontrol2} \mathrm{s.t.} \quad &    \bm{\dot{x}}=\bm{f}(\bm{x},\,\bm{u}) \\
      \label{eq:optimalcontrol3}               \quad &    \bm{x}(0)=\bm{x}_i, \quad \bm{x}(t_f)=\bm{x}_f \\
      \label{eq:optimalcontrol4}               \quad &    \bm{x}\in \bm{X}, \qquad \ \ \bm{u}\in \bm{U}
  \end{align}
  \label{eq:optimalcontrol}
\end{subequations}
\noindent\!\!where $\bm{x}$ and $\bm{u}$ are trajectory functions of time $t\in [0,\, t_f]$, i.e., we can write $\bm{x}(t)$ and $\bm{u}(t)$ explicitly. Equation (\ref{eq:optimalcontrol2}) represents vehicle dynamics constraints, $\bm{x}_i$ and $\bm{x}_f$ are vehicle initial and final states, respectively. However, at some times, the vehicle needs to stop at a goal set instead of an exact goal state, i.e. $\bm{x}(t_f)\in \bm{X}_f$, e.g., the vehicle stops inside a parking spot other than a perfect location. The cost function $g(\bm{x}_i,\,\bm{x}_f,\,t_f)$ depends only on the initial state $\bm{x}_i$, final state $\bm{x}_f$ and the completion time $t_f$, while $L(\bm{x},\,\bm{u})$ describes some objective along the trajectory such as the energy consumption and $0\le\beta\le1$ is the weight. A particular performance index is to minimize the weight sum of completion time and energy consumption:
\begin{equation}
    t_f + \int_0^{t_f} \upsilon^2 \mathrm{d}t   
    \label{eq:objective}
\end{equation}

The problem (\ref{eq:optimalcontrol}) is a traditional optimal control problem that can handle any nonlinear dynamics and (piecewise) continuous trajectory constraints. Adding point-point distance constraints between circles \cite{robinson2018efficient} or area criterion constraints between rectangles \cite{li2015unified}, a motion planning model can be obtained. However, neither collision avoidance methods can be used to general fast high-precision motion planning problems. Adding a standard linear programming collision avoidance problem as an embedded constraint to (\ref{eq:optimalcontrol}), the BOMP model is proposed:  
\begin{subequations}
  \begin{align}
      \min \quad &    \beta g(\bm{x}_i,\,\bm{x}_f,\,t_f) + (1-\beta)\int_0^{t_f} L(\bm{x},\,\bm{u}) \mathrm{d}t \nonumber \tag{4} \\ 
      \mathrm{s.t.} \quad &    \bm{\dot{x}}=\bm{f}(\bm{x},\,\bm{u})  \\
         \quad &   \bm{x}(0)=\bm{x}_i, \quad \bm{x}(t_f)=\bm{x}_f  \\
         \quad &   \bm{x}\in \bm{X}, \qquad \ \ \bm{u}\in \bm{U}  \\
         \label{eq:gframwork1}       \quad &    J\ge\delta>0 \\
         \label{eq:gframwork2}        \quad &    J = \min \  \bm{c}^\textrm{T}\bm{y} \\
         \quad & \qquad \mathrm{s.t.} \ \    \bm{Ay} = \bm{b}, \quad \bm{y}\ge\bm{0}. \nonumber 
  \end{align}
  \label{eq:gframwork}
\end{subequations}
\!\!where $\bm{c}\in\Re^n$ and $\bm{b}\in\Re^m$ are constant vectors, $\bm{y}\in\Re^n$ is an extra added time dependent variable, and $\bm{A}\in\Re^{m\times n}$ depends on vehicle state variable $\bm{x}$. Besides, $J$ is the optimal value of the linear programming problem, it's a time-dependent scalar variable and $\delta$ is a constant scalar. Therefore, the linear programming (\ref{eq:gframwork2}) represents constraints on vehicle state trajectory. Suppose that the linear programming (\ref{eq:gframwork2}) represents a distance function between vehicle and the surrounding obstacles, and $J>0$ denotes collision free.  Then, the constraints (\ref{eq:gframwork1}) and (\ref{eq:gframwork2}) denote that the vehicle configuration $\bm{q}$ is in the $\delta$-interior of collision-free space  \cite{karaman2011sampling}, that is:
\begin{equation}
  \begin{aligned}
      & J\ge\delta>0 \\
      & J = \min \  \bm{c}^\textrm{T}\bm{y}  \qquad\qquad \Leftrightarrow \quad \bm{q}\in\mathrm{int}_\delta(\mathcal{\bm{C}}_{\mathrm{free}}) \\
      & \mathrm{s.t.} \ \    \bm{Ay} = \bm{b}, \ \bm{y}\ge\bm{0}. 
  \end{aligned}
  \label{eq:deltaequality}
\end{equation}

As we can see, the optimal control model (the upper-level problem) (\ref{eq:gframwork}) contains an embedded optimization problem (the lower-level problem) as its constraint, and to our knowledge this model has never appeared in the literature before. Though this is a bilevel optimal control problem, the complexity and convergence analysises can be seen indirectly from the bilevel optimization problems as stated in section \ref{sec:problemanalysis}. In section \ref{sec:Jfunction}, we present the $J_2$-function collision avoidance method which checks collision between convex polyhedrons in any dimensional space, and can be applied to many domains. However, this paper focuses on fast high-precision autonomous parking problems, the general applicability of the BOMP model will be discussed in future works.

\section{Collision Avoidance}
\label{sec:Jfunction}
The performances of robot collision-free motion planning algorithms depend highly on the geometry collision avoidance methods used between robot and the environment. In real-world, robot and environment geometries are always sophisticated such that the approximation techniques are always utilized to simplify the collision avoidance problem. Polyhedron or sphere approximation in 3D environment and polygon or circle approximation in 2D environment are the conventional methods. In this section, point-point distance constraints between spheres (or circles), area criterion constraints between convex polygons and the $J_2$-function are summarized first. The $J_2$-function represents one collision avoidance method between convex polyhedrons in any dimensional spaces. The in-depth geometric interpretation of the artificial variables in the $J_2$-function is presented. Then the MJ function and the APMJ function are proposed according to the geometric properties. 
\subsection{Basic Collision Avoidance Constraints}
\label{sec:straightforward}
Suppose robot $\bm{B}$ is approximated by a sphere or circle of radius $r_b$ that covers it, and obstacle $\bm{A}$ is approximated by a sphere or circle of radius $r_a$. The centers of the approximate objects are $\bm{p}_b$ and $\bm{p}_a$. Then the geometry collision-free constraint between $\bm{A}$ and $\bm{B}$ is approximated as:
\begin{equation}
    d(\bm{p}_a,\,\bm{p}_b) - r_a - r_b \ge \delta >0
    \label{eq:circlecollisionfree}
\end{equation}
where $d(\bm{p}_a,\,\bm{p}_b)$ denotes the Euclidean distance between the two points. Usually, the robot and the obstacles are approximated by several overlapping spheres or circles to increase approximate precision, in this case, the geometry collision-free constraints are still straightforward.

Suppose plane robot $\bm{B}$ and plane obstacle $\bm{A}$ are approximated by convex polygons with $n_b$ vertexes and $n_a$ vertexes, respectively. The vertexes coordinates of the polygons are $\{\bm{p}_i\}, i=1,2,\cdots,n_b$ and $\{\bm{p}_j\}, j=1,2,\cdots,n_a$, respectively. Then the area criterion can describes the geometry collision-free constraints between $\bm{A}$ and $\bm{B}$:   
\begin{equation}
    \begin{aligned}
        & \sum_{j=1}^{n_a-1} S_{\triangle \bm{p}_i\bm{p}_j\bm{p}_{j+1}} + S_{\triangle \bm{p}_i\bm{p}_{n_a}\bm{p}_1} - S_{\bm{A}} \ge \delta >0,\,\, i=1,2,\cdots,n_b\\ 
        & \sum_{i=1}^{n_b-1} S_{\triangle \bm{p}_j\bm{p}_i\bm{p}_{i+1}} + S_{\triangle \bm{p}_j\bm{p}_{n_b}\bm{p}_1} - S_{\bm{B}} \ge \delta >0,\,\, j=1,2,\cdots,n_a
    \end{aligned}
    \label{eq:circlecollisionfree}
\end{equation}
where $S_{\triangle}$ is the area of a triangle, $S_{\bm{A}}$ and $S_{\bm{B}}$ are the areas of the two approximate polygons. Moreover, for polyhedrons in 3D environment, the volume criterion can derive the collision avoidance constraints.

As we can see, for a pair of polygons, there are $n_a+n_b$ nonlinear constraints in (\ref{eq:circlecollisionfree}), and this number is usually larger than the circle approximation method. However, in theory, a complex geometry can be incorporated by several polyhedrons or polygons in any precision and low overlapping rate with fewer approximate objects than the sphere or circle approximation method. And to achieve planning and control in scenarios with  long and thin obstacles, the polyhedron or polygon approximation is more efficient. Therefore, the convex polyhedron or polygon approximation method will play an important role in complex and high-precision robot motion planning problems.
\subsection{Definition and Basic Properties of the $J_2$-function}
\label{sec:J function}
Given two convex polyhedron objects that are represented by the convex hulls of two point sets $\bm{A}=\{\bm{a}_i\}, i=1,2,\cdots,n_1$ and $\bm{B}=\{\bm{b}_j\}, j=1,2,\cdots,n_2$, respectively, i.e., $\mathrm{conv}\bm{A}$ and $\mathrm{conv}\bm{B}$. Where $\bm{a}_i \in \Re^m$ and $\bm{b}_j \in \Re^m$ are point coordinates, $m$ is the dimensionality, $n_1$ and $n_2$ are the point numbers. Then the $J_2$-function checks whether the two objects $\mathrm{conv}\bm{A}$ and $\mathrm{conv}\bm{B}$ are colliding or not, and its form is:
\begin{subequations}
  \begin{align}
       J_2(\bm{A},\,&\bm{B}) = \min \sum_{k=1}^{m+2}z_k  \nonumber \tag{8} \\
      \label{eq:Jfunction1} \mathrm{s.t.} \quad & \bm{Ax} - \bm{By} + \bm{z} = \bm{0}, \\
            \quad & \sum_{i=1}^{n_1}x_i + z_{m+1} = 1, \quad \sum_{j=1}^{n_2}y_j + z_{m+2} = 1, \\
            \quad & \bm{x}\ge\bm{0},\ \bm{y}\ge\bm{0}, \ \bm{z}\ge\bm{0} \\
       \label{eq:Jfunction4}     \quad & z_{m+1}\ge0,\ z_{m+2}\ge0 
  \end{align}
  \label{eq:Jfunction}
\end{subequations}
\noindent\!\!where $\bm{x}=[x_1 \ x_2 \ \cdots \ x_{n_1}]^\mathrm{T}$, $\bm{y}=[y_1 \ y_2 \ \cdots \ y_{n_2}]^\mathrm{T}$, and $\bm{z}=[z_1 \ z_2 \ \cdots \ z_m]^\mathrm{T}$. The vector $\bm{z}$ and scalars $z_{m+1},\, z_{m+2}$ are called artificial variables. In constraint (\ref{eq:Jfunction1}), the symbols $\bm{A}$ and $\bm{B}$ are actually matrixes composed of the points coordinates, i.e., $\bm{A}=[\bm{a}_1 \ \bm{a}_2 \ \cdots \ \bm{a}_{n_1}]$ and $\bm{B}=[\bm{b}_1 \ \bm{b}_2 \ \cdots \ \bm{b}_{n_2}]$, so in this paper, the notations of point set and matrix are not distinguished explicitly, and the true meaning can be seen directly and easily through context. Besides, when referring to a point set of one object, it is actually the convex hull of it is used to characterize that object. Throughout this paper, vector inequality is to be understood element-wise.

The $J_2$-function has the following basic properties:
\begin{enumerate}[leftmargin=4mm,labelsep=0.2em,topsep=2pt,itemsep=0pt,parsep=1pt,label=(\alph*)]
    \item the feasible region (\ref{eq:Jfunction1})-(\ref{eq:Jfunction4}) is nonempty; 
    \item the range of the optimal value is $0\le J_2 \le2$; 
    \item $J_2>0$ is equivalent to $\mathrm{conv}\bm{A}\cap \mathrm{conv}\bm{B}=\emptyset$, i.e., collision free between the objects; 
    \item it provides a concept of pseudodistance between convex polyhedrons and is pseudomonotonic with respect to the Euclidean distance. 
\end{enumerate}

\subsection{Geometric Interpretation of the $J_2$-function}
\label{sec:Jfunctiongeometric}
\begin{figure}[t]
    \centering
    \includegraphics[width=0.33\textwidth]{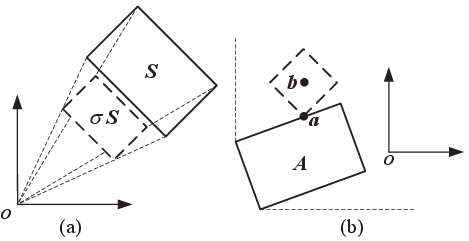}
    \caption{The geometric meaning of the artificial variables $\bm{z}$, $z_{m+1}$ and $z_{m+2}$. (a) The shrinking model of convex polygon $\bm{S}$ with respect to the origin and $\sigma$ is the fractional shrink. $z_{m+1}$ and $z_{m+2}$ are related to the fractional shrinks of objects $\bm{A}$ and $\bm{B}$. (b) The $l_1$ distance model between a point and a convex polygon and $\bm{z}$ denotes the vector from point $\bm{a}$ to $\bm{b}$.}
    \label{fig:geometricproof}
\end{figure}

Suppose $\sigma_A=\sum_{i=1}^{n_1}x_i$ and $\sigma_B=\sum_{j=1}^{n_2}y_j$, then $0\le\sigma_A,\sigma_B\le1$ and they denote the fractional shrinks of objects $\bm{A}$ and $\bm{B}$ with respect to the origin respectively, as shown in Fig. \ref{fig:geometricproof}(a). In this way, the meaning of the artificial variables $z_{m+1}$ and $z_{m+2}$ can be seen directly:
\begin{equation}
    z_{m+1} = 1-\sigma_A,\ z_{m+2} = 1-\sigma_B
    \label{eq:artificalmeaning}
\end{equation}

From the definition of $\sigma_B$, we can get $\bm{By}=\sigma_B \bm{p}_B$ where $\bm{p}_B\in \mathrm{conv}\bm{B}$. So in order to simplify the understanding of the artificial vector $\bm{z}$, a point set $\bm{A}$ and a point $\bm{b}$ are considered in the $J_2$-function. We first make the assumption that $\sigma_A=\sigma_B=1$ (or $z_{m+1}=z_{m+2}=0$) and concentrate on the meaning of vector $\bm{z}$. Since $\bm{z}\ge\bm{0}$, from the constraint (\ref{eq:Jfunction1}), the basic feasible domain is $\bm{b}\in \mathrm{conv}\bm{A}+\Re_+^m$ where $\Re_+^m$ denotes the nonnegative orthant. We then let $\mathrm{conv}\bm{A}_+=\mathrm{conv}\bm{A}+\Re_+^m$ and make the assumptions that $\bm{b}\in \mathrm{conv}\bm{A}_+$ and $\bm{b}\cap\mathrm{conv}\bm{A}=\emptyset$ as the case shown in Fig. \ref{fig:geometricproof}(b) (the exception cases will be explained later). Combined with the objective of the $J_2$-function, it can be seen that the $l_1$ norm of the optimal solution $\bm{z}^*$ equals to the optimal value $J_2(\bm{A},\bm{b})$, that is, the artificial vector $\bm{z}$ measures the $l_1$ distance between two convex polyhedrons. And the following two geometric problems (\ref{eq:lemma1}) and (\ref{eq:lemma2}) can be obtained. 

Consider a point set $\bm{A}$, a point $\bm{b}$ and suppose that $\sigma_A=\sigma_B=1$ and $\bm{b}\in \mathrm{conv}\bm{A}_+$, then the $J_2$-function is equivalent to the following geometric problem:  
\begin{equation}
    J_2(\bm{A},\bm{b})=\min_{\bm{S}_c\cap \mathrm{conv}\bm{A}\ne \emptyset,\ c\ge0} c
    \label{eq:lemma1}
\end{equation}
where $\bm{S}_c$ denotes the level set of the $l_1$ distance function with respect to point $\bm{b}$.
\begin{equation*}
    \bm{S}_c=\{\bm{p}\in\Re^m \ | \ c=\Vert \bm{p}-\bm{b} \Vert_1\}
\end{equation*}

Consider point sets $\bm{A}$ and $\bm{B}$, and suppose that $\sigma_A=\sigma_B=1$ and $\mathrm{conv}\bm{B}\cap \mathrm{conv}\bm{A}_+\ne\emptyset$, then the $J_2$-function is equivalent to the following geometric problem:  
\begin{equation}
    J_2(\bm{A},\bm{B})=\min_{\bm{S}_c\cap \mathrm{conv}\bm{A}\ne \emptyset,\ c\ge0} c
    \label{eq:lemma2}
\end{equation}
where $\bm{S}_c$ denotes the level set of the $l_1$ distance function with respect to object $\bm{B}$.
\begin{equation*}
    \bm{S}_c=\{\bm{p}\in\Re^m \ | \ c=\min_{\bm{p}\in \mathrm{conv}\bm{A}_+,\bm{b}\in\mathrm{conv}\bm{B}}\Vert \bm{p}-\bm{b} \Vert_1\}
\end{equation*}
\begin{figure}
    \centering
    \includegraphics[width=0.27\textwidth]{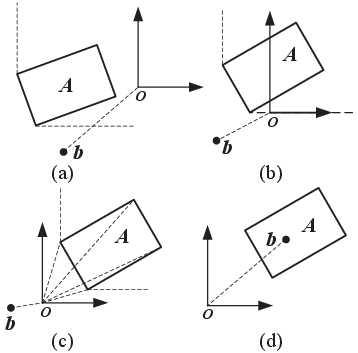}
    \caption{Some special relationships of a point $\bm{b}$ with respect to a convex polygon $\bm{A}$ in the plane.}
    \vspace{-1.0ex}
    \label{fig:geometriccases}
\end{figure}

As the geometric meaning of $\bm{z}$, $z_{m+1}$ and $z_{m+2}$ are clear, we now discuss how the $J_2$-function works when $\bm{b}\notin \mathrm{conv}\bm{A}_+$. This case can be categorized into three subcases as shown in Figs. \ref{fig:geometriccases}(a)-(c): (a) $\bm{0}\in \mathrm{int} \, \mathrm{conv}\bm{A}_+$ and $\bm{b}\notin \mathrm{conv}\bm{A}_+$; (b) $\bm{0}\in \mathrm{bd} \, \mathrm{conv}\bm{A}_+$ and $\bm{b}\notin \mathrm{conv}\bm{A}_+$; and (c) $\bm{0}\notin \mathrm{conv}\bm{A}_+$ and $\bm{b}\notin \mathrm{conv}\bm{A}_+$. Where $\mathrm{int} \, \mathrm{conv}\bm{A}_+$ and $\mathrm{bd} \, \mathrm{conv}\bm{A}_+$ denote the interior and the boundary of $\mathrm{conv}\bm{A}_+$, respectively. In subcase (a), $\exists\sigma_B>0,\ \sigma_B \bm{b}\in \mathrm{conv}\bm{A}_+$ such that the linear programming can be solved. In subcase (b), $\sigma_B$ must equals zero to make the constraint feasible. While in subcase (c), the optimal solution can be obtained intuitively: $\sigma_A^*=0,\ \sigma_B^*=0$ and $J_2=2$. In the end, we point out that the optimal value $J_2=0$ only occurs when $\bm{b}\in \mathrm{conv}\ \bm{A}$ or $\mathrm{conv}\ \bm{A} \cap \mathrm{conv}\ \bm{B}\neq\emptyset$. 
\subsection{Modified $J_2$-function}
The geometric meaning of the artificial variables states that the $J_2$-function achieves to maximize the shrinks $\sigma_A$, $\sigma_B$ as well as minimize the $l_1$ distance between $\sigma_A\mathrm{conv}\bm{A}$ and $\sigma_B\mathrm{conv}\bm{B}$ with constraint $\sigma_B\mathrm{conv}\bm{B}\cap (\sigma_A\mathrm{conv}\bm{A}+\Re_+^m)\ne \emptyset$. In this subsection, we will show the $J_2$-function can be realized by only maximizing $\sigma_B$ and get the geometric problem (\ref{eq:mjgeometric}).

Given point sets $\bm{A}$ and $\bm{B}$, and suppose that $\bm{0}\in\mathrm{int}\,\mathrm{conv}\,\bm{A}$, then the $J_2$-function is equivalent to the geometric problem:  
\begin{equation}
    J_2(\bm{A},\bm{B})=\min_{\sigma_B\mathrm{conv}\bm{B}\cap \mathrm{conv}\bm{A} \ne\emptyset,\ 0\le\sigma_B\le1} 1-\sigma_B
    \label{eq:mjgeometric}
\end{equation}

\begin{figure}
    \centering
    \includegraphics{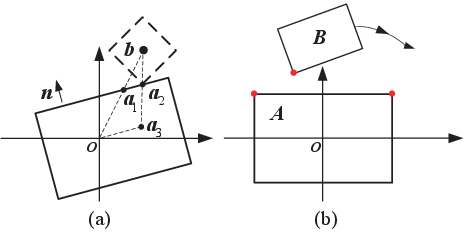}
    \caption{(a) The significance comparison between the shrink $\sigma_B$ and the $l_1$ distance. (b) The active points (red points) can characterize the collision behavior at the instant.}
    \label{fig:mj}
\end{figure}

When $\bm{0}\in\mathrm{int}\,\mathrm{conv}\,\bm{A}$ and $\bm{b}\notin \mathrm{conv}\bm{A}_+$, the correctness of problem (\ref{eq:mjgeometric}) can be seen directly from the geometric analysis in Section \ref{sec:Jfunctiongeometric}. Therefore, we suppose $\bm{0}\in\mathrm{int}\,\mathrm{conv}\,\bm{A}$ and $\bm{b}\in \mathrm{conv}\bm{A}_+$, and a rectangle $\bm{A}$ and a point $\bm{b}$ as shown in Fig. \ref{fig:mj}(a) are used to illustrate intuitively of the proof. Since $\bm{0}\in\mathrm{int}\,\mathrm{conv}\,\bm{A}$, $\forall \bm{b}\in \mathrm{conv}\bm{A}_+, \exists0\le\sigma_B\le1, \sigma_B \bm{b}\cap\mathrm{conv}\bm{A}\ne\emptyset$. And if $\bm{b}\notin\mathrm{conv}\bm{A}$, then as $\sigma_B$ approaches zero, $\sigma_B \bm{b}$ intersects the boundary of $\mathrm{conv}\bm{A}$ at first. Besides, if $\sigma_A$ gets small, then $\sigma_B$ will become smaller to make $\sigma_B \bm{b}$ intersect the boundary of $\sigma_A\mathrm{conv}\bm{A}$. As a result, the optimal value $\sigma_A^*$ equals one (the same result can be obtained from the view of the $l_1$ distance between point $\bm{b}$ and object $\bm{A}$). Now, only the shrink $\sigma_B$ and the $l_1$ distance influence the objective value of the $J_2$-function. Suppose $\bm{b}\notin\mathrm{conv}\bm{A}$, then the $l_1$ distance gets small as $\sigma_B$ approaches zero, so the $J_2$-function achieves a balance between them. To compare the significance of these two factors in the $J_2$-function, they are considered independently. Let $\sigma_B^\prime$ denotes the optimal solution of $\max_{\sigma_B \bm{b}\cap\mathrm{conv}\bm{A}\ne\emptyset} \sigma_B$ and let $\bm{a}_1=\sigma_B^\prime \bm{b}$. Let $c^\prime$ denote the optimal value of (\ref{eq:lemma1}), $\bm{a}_2=\bm{S}_{c^\prime}\cap \mathrm{conv}\bm{A}$, $\bm{n}$ denote the normal of the surface from $\mathrm{conv}\,\bm{A}$ that intersects with $\bm{S}_{c^\prime}$, and $\bm{a}_3$ denote the intersection point of the line $l_{ba_2}$ and the subspace whose normal is $\bm{n}$. The notations $\bm{a}_1,\,\bm{a}_2,\,\bm{a}_3$ and $\bm{n}$ are shown in Fig. \ref{fig:mj}(a), then:
\begin{equation}
    c^\prime=\Vert \bm{b}-\bm{a}_2 \Vert, \ \  1-\sigma_B^\prime=\frac{\Vert \bm{b}-\bm{a}_1 \Vert}{\Vert \bm{b} \Vert}=\frac{\Vert \bm{b}-\bm{a}_2 \Vert}{\Vert \bm{b}-\bm{a}_3 \Vert}<1.
    \label{eq:jmjproof}
\end{equation}
It can be seen that $c^\prime$ is much bigger than $1-\sigma_B^\prime$, such that $\sigma_B^\prime$ plays the leading role in the $J_2$-function. Therefore, $\sigma_B^\prime=\sigma_B^*=1-J_2(\bm{A},\bm{b})$ and the problem (\ref{eq:mjgeometric}) can be obtained.

According to the geometric problem (\ref{eq:mjgeometric}), if $\bm{0}\in\mathrm{int}\,\mathrm{conv}\,\bm{A}$, the MJ linear programming is:
\begin{subequations}
  \begin{align}
    J_2(\bm{A},&\bm{B}) = \min \ 1-\sum_{j=1}^{n_2}y_j \nonumber \tag{14} \\
    \label{eq:Jimprovement2} \mathrm{s.t.} \quad & \bm{Ax} - \bm{By} = \bm{0}, \\
                                 \quad & \sum_{i=1}^{n_1}x_i = 1,\quad \sum_{j=1}^{n_2}y_j \le 1, \\
    \label{eq:Jimprovement4}      \quad & \bm{x}\ge\bm{0},\ \bm{y}\ge\bm{0} 
  \end{align}
  \label{eq:Jimprovement}
\end{subequations}
\!\!\!And its compact form:
\begin{equation}
  \begin{aligned}
      J_2(\bm{A},\bm{B}&) = \min \ 1+\bm{c}^\mathrm{T}\bm{p} \\
      \mathrm{s.t.} \quad & \bm{Qp}=\bm{b}, \quad 1+\bm{c}^\mathrm{T}\bm{p}\ge0, \quad \bm{p}\ge\bm{0}\\
  \end{aligned}
  \label{eq:Jimprovementcompact}
\end{equation}
where
\begin{equation*}
    \begin{aligned}
        \bm{p}=&\left[\begin{array}{c} \bm{x} \\ \bm{y} \\ \end{array}\right], \ \bm{c}=\left[\begin{array}{c} \bm{0}_{n_1\times 1} \\ -\bm{1}_{n_2\times 1} \\ \end{array}\right], \
        \bm{b}=\left[\begin{array}{c} \bm{0}_{m\times 1} \\ 1 \\ \end{array}\right], \\ 
        \bm{Q}=&\left[\begin{array}{cc} \bm{A} & -\bm{B} \\ \bm{1}_{1\times n_1} & \bm{0}_{1\times n_2} \\ \end{array}\right].
    \end{aligned}
\end{equation*}

The MJ function reduces $m+2$ variables and increases computation speed. The optimal value is $0\le J_2 \le1$, while $J_2=1$ corresponds to the case that $\bm{B}$ is at infinity.

\subsection{Active Points based Modified $J_2$-function}
In fact, the collision between two relatively moving polyhedrons always arises between the closest components, face/vertex, vertex/face or edge/edge \cite{lozano1983spatial}, so the collision avoidance is equivalent to the closest components are collision free. According to this fact, if the closest components can be determined, the instantaneous geometric representation of each object can be simplified for collision avoidance problem. In this subsection, we make the assumption that the given two objects are not colliding such that the concept of closest component is meaningful. Fig. \ref{fig:mj}(b) shows two relatively moving plane polygons $\bm{A}$ and $\bm{B}$ that statisfy the assumption. Then at the instant, each polygon can be characterized by the red points to avoid collision. In this paper, these points are called active points and they correspond to the nonzero elements of the optimal solution $\bm{x}^*,\,\bm{y}^*$ in the $J_2$-function: 
\begin{equation}
    \bm{A}_a=\{\bm{a}_i \ | \ x_i^* \neq 0\}, \ \bm{B}_a=\{\bm{b}_j \ | \ y_j^* \neq 0 \}
    \label{eq:activepoints}
\end{equation}

Using $\bm{A}_a$ and $\bm{B}_a$, an APMJ function can be obtained: 
\begin{equation}
  \left\{
      \begin{aligned}
          &J_2(\bm{A}_a,\bm{B}_a) =\ 1-\sum_{j=1}^{n_{a2}}y_j \\
          &\bm{A}_a\bm{x} - \bm{B}_a\bm{y} =\bm{0}, \quad \sum_{i=1}^{n_{a1}}x_i = 1, \\
          &\bm{x}\ge\bm{0},\quad \bm{y}\ge\bm{0}
      \end{aligned}
  \right.
  \label{eq:japmj}
\end{equation}
where ${n_{a1}}$ and ${n_{a2}}$ are the active points numbers. 

Note that the APMJ function is not an optimization problem, it is just a constrained system of linear equations, besides, the cardinalities of point sets $\bm{A}_a, \bm{B}_a$ are very low. Therefore, the APMJ function reduces the computation time dramatically. However, given two objects arbitrarily, the active points between them are not transparent. Moreover, in robot motion planning problem, the robot motion trajectory is not known first, let alone the active points. As a consequence, the APMJ function can not be used straightway, some techniques have to be utilized to pick out the active points first. In Section \ref{sec:bomp}, a two-stage BOMP algorithm is proposed. The MJ function at the initial stage finds an approximate optimal trajectory, and then the active points along the trajectory are picked out to feed into the APMJ function to construct the final stage algorithm. The related theory, and the two-stage algorithm details will be discussed in the following sections.

\section{Solution of the BOMP Model}
\label{sec:bomp}
\subsection{$J_2$-function Based BOMP Model}
Consider a convex polyhedron robot $\bm{B}$ moves around a static convex polyhedron obstacle $\bm{A}$ from an initial state to a final state. And suppose that the purpose is to generate an optimal smooth collision-free trajectory. According to properties (c) and (d) of the $J_2$-function, a new constraint $J_2(\bm{A},\bm{B})\ge\delta>0$ and the MJ function replace the constraints (\ref{eq:gframwork1}) and (\ref{eq:gframwork2}), an exact $J_2$-function based BOMP model is obtained:
\begin{subequations}
  \begin{align}
      \min \quad &    \beta g(\bm{x}_i,\,\bm{x}_f,\,t_f) + (1-\beta)\int_0^{t_f} L(\bm{x},\,\bm{u}) \mathrm{d}t \tag{18}\\ 
      \mathrm{s.t.} \quad &    \bm{\dot{x}}=\bm{f}(\bm{x},\,\bm{u})  \\
         \quad &   \bm{x}(0)=\bm{x}_i, \quad \bm{x}(t_f)=\bm{x}_f  \\
         \quad &   \bm{x}\in \bm{X}, \qquad \ \ \bm{u}\in \bm{U}  \\
         \quad &    J_2(\bm{A},\,\bm{B})\ge\delta>0  \label{eq:framwork5}\\
         \quad &    J_2(\bm{A},\,\bm{B}) = \min \  1+\bm{c}^\textrm{T}\bm{p} \label{eq:framwork6}\\
         \quad & \, \mathrm{s.t.} \quad \,   \bm{Qp} = \bm{b}, \quad 1+\bm{c}^\textrm{T}\bm{p}\ge0, \quad \bm{p}\ge\bm{0}.\nonumber 
  \end{align}
  \label{eq:framwork}
\end{subequations}
\noindent\!\!\!where $\bm{A}$ and $\bm{B}$ are point sets described in the obstacle coordinate system. Therefore, $\bm{A}$ is constant, and $\bm{B}$ is a function of robot state $\bm{x}$. More precisely, $\bm{B}$,\,$\bm{Q}$,\,$\bm{p}$ are trajectory functions of time $t$. Therefore, the constraint $J_2(\bm{A},\bm{B})\ge\delta>0$ shows robot motion is collision-free.
\subsection{Model Solution}
\label{sec:bilevel}
Given the problematic nature of the BOMP model, it is helpful to reduce the overall bilevel optimal control problem to a traditional single level optimal control problem. According to the optimality conditions of linear programming \cite{wright1999numerical} and noticing the constraint (\ref{eq:framwork5}), the constraint $1+\bm{c}^\textrm{T}\bm{p}\ge0$ is not active. So in the lower level optimization problem (\ref{eq:framwork6}), there are Lagrange multipliers $\bm{\lambda}\in\Re^{n_1+n_2} \ge \bm{0}$ corresponding to constraint $\bm{p}\ge\bm{0}$ and multipliers $\bm{v}\in\Re^{m+1}$ corresponding to constraint $\bm{Qp} = \bm{b}$. $\bm{\lambda}$ and $\bm{v}$ are trajectory functions of time $t$, then the continuous MAKKT form of the lower problem is: 
\begin{subnumcases}{}
    \label{eq:lowerMAKKT1} \bm{Qp} = \bm{b},\quad \bm{p}\ge\bm{0} \\
    \label{eq:lowerMAKKT2} \Vert \bm{c}-\bm{\lambda}+\bm{Q}^\mathrm{T}\bm{v} \Vert \le \sqrt{\epsilon},\quad \bm{\lambda}^\mathrm{T} \bm{p} \le \epsilon,\quad \bm{\lambda}\ge\bm{0}  
\end{subnumcases}
where (\ref{eq:lowerMAKKT1}) are feasible conditions, (\ref{eq:lowerMAKKT2}) are equilibrium and complementary conditions and $\epsilon>0$ is the approximate relaxation. A point $\bm{p}$ statisfing these constraints is called an $\epsilon$-MAKKT point (or $\epsilon$ approximate optimal point), and note that when $\epsilon=0$ it's actually the KKT point (or optimal point).

\begin{figure}[t]
  \vspace{-3ex}
  \centering
  \includegraphics{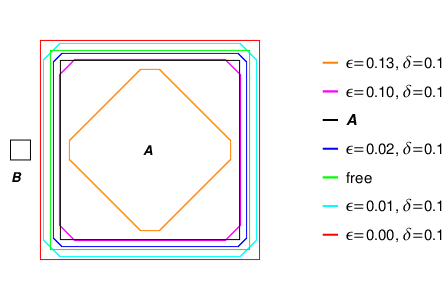}
  \vspace{-3ex}
    \caption{The square $\bm{B}$ is free to translate with fixed orientation, thus its configuration can be described by the coordinate of its center point and the configuration space $\mathcal{\bm{C}}$ is the plane. The square $\bm{A}$ is the obstacle and the coordinate system is origined at its center. The label "$\bm{A}$" denotes the obstacle, the label "free" denotes the boundary of obstacle free configuration space $\mathcal{\bm{C}}_{\mathrm{free}}$, the label "$\epsilon=0.00, \delta=0.1$" denotes the $\mathrm{int}_\delta(\mathcal{\bm{X}}_{\mathrm{free}})$ and the other labels denote $\epsilon$-$\mathrm{int}_\delta(\mathcal{\bm{C}})$. As we can see, $\epsilon$-$\mathrm{int}_\delta(\mathcal{\bm{C}})$ converges to $\mathrm{int}_\delta(\mathcal{\bm{C}}_{\mathrm{free}})$ as $\epsilon$ converges to zero.} 
  \label{fig:cspaces}
\end{figure}

Since in the BOMP model the optimal value $J_2(\bm{A},\bm{B})$ of the lower level optimization problem is also restricted, the relation between the optimal value and the approximate optimal value also needs to be considered. The \textbf{Theorem 3.5} in \cite{dutta2013approximate} helps the analysis, that is, at an $\epsilon$-MAKKT point, the objective value $1+\bm{c}^\mathrm{T}\bm{p}$ is at least $\epsilon$ larger than $J_2(\bm{A},\bm{B})$. Consequently, the converted single level optimal motion planning problem for (\ref{eq:framwork}) can be formulated as follow: 
\begin{subequations}
  \begin{align}
      \min \quad &   \beta g(\bm{x}_i,\bm{x}_f,t_f) + (1-\beta)\int_0^{t_f} L(\bm{x},\bm{u}) \mathrm{d}t \nonumber \tag{20} \\  
      \mathrm{s.t.} \quad &    \bm{\dot{x}}=\bm{f}(\bm{x},\bm{u}) \label{eq:makktframwork4}\\
      \quad &    \bm{x}(0)=\bm{x}_i, \quad \bm{x}(t_f)=\bm{x}_f  \\
      \quad &    \bm{x}\in \bm{X}, \qquad \ \ \bm{u}\in \bm{U} \\
      \quad & 1+\bm{c}^\mathrm{T}\bm{p}\ge \epsilon+\delta \label{eq:makktframwork1} \\
      \quad &    \bm{Qp} = \bm{b},\quad \bm{p}\ge\bm{0},\quad \bm{\lambda}\ge\bm{0}  \label{eq:makktframwork2} \\
      \quad &    \Vert \bm{c}-\bm{\lambda}+\bm{Q}^\mathrm{T}\bm{v} \Vert \le \sqrt{\epsilon},\quad \bm{\lambda}^\mathrm{T} \bm{p} \le \epsilon  \label{eq:makktframwork3}  
  \end{align}
  \label{eq:makktframwork}
\end{subequations}
\indent\!\!The relationship between the MAKKT and the exact KKT optimality conditions is as follow: if a sequence of points $\{\bm{p}_k\}$  converge to a point $\bar{\bm{p}}$ where the Mangasarian Fromovitz constraint qualification is also satisfied, each point $\bm{p}_k$ is an $\epsilon_k$-MAKKT point, and $\epsilon_k \rightarrow 0$ as $k \rightarrow \infty$, then the point $\bm{\bar{p}}$ is a KKT point. A robot state satisfying constraints (\ref{eq:framwork5}) and (\ref{eq:framwork6}) means that the configuration is in the $\delta$-interior of free space, i.e., $\bm{q}\in\mathrm{int}_\delta(\mathcal{\bm{C}}_{\mathrm{free}})$ \cite{karaman2011sampling}. Similarly, we define a state $\bm{x}\in\bm{X}$ satisfying the constraints (\ref{eq:makktframwork1})-(\ref{eq:makktframwork3}) means that the configuration is an $\epsilon$-$\delta$-interior configuration. And the $\epsilon$-$\delta$-interior of $\bm{\mathcal{C}}$, denoted as $\epsilon$-$\mathrm{int}_\delta(\bm{\mathcal{C}})$, is defined as the collection of all $\epsilon$-$\delta$-interior configurations. According to the relationship between the MAKKT and the exact KKT optimality conditions, $\epsilon$-$\mathrm{int}_\delta(\bm{\mathcal{C}})$ converges to $\mathrm{int}_\delta(\bm{\mathcal{C}}_{\mathrm{free}})$ as $\epsilon$ converges to zero (seen in Fig. \ref{fig:cspaces}).

Therefore, in order to solve problem (\ref{eq:framwork}), a decreasing sequence $\epsilon$ should be applied to problem (\ref{eq:makktframwork}) and a sequence of approximate optimal trajectory is solved to converge to the exact optimal trajectory. When $\epsilon$ is large, problem (\ref{eq:makktframwork}) is easy to solve an $\epsilon$ optimal trajectory. Then decrease $\epsilon$ and choose the previous trajectory as an initial guess to solve a new trajectory until $\epsilon$ converges zero. However, as $\epsilon$ approaches zero, the nonconvexities and combinatorial property in the MAKKT conditions become highlighted, then the problem becomes hard to solve or nonconvergent. Under this circumstance, a new convergence technique needs to be sought. The APMJ function settles this issue, and why it works is as follows. 
\begin{figure}[t]
  \centering
  \includegraphics{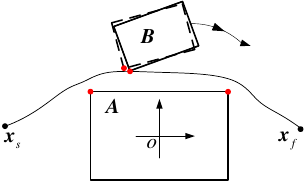}
  \caption{The reason why the APMJ function works. The solid line is an approximate optimal trajectory for object $\bm{B}$ moving from $\bm{x}_s$ to $\bm{x}_f$, and a small variation does not change the active points between the two objects.}
  \label{fig:apvariant}
\end{figure}

As $\epsilon$ approaches zero, it will solve a collision-free approximate optimal trajectory first, and the exact optimal solution is a small variation of it as shown in Fig. \ref{fig:apvariant}, where the solid line and the dashed box are the approximate trajectory and an instant optimal location, respectively. As we can see, the small variation does not change the active points between the objects, so the active-points from the approximate trajectory can characterize the collision behavior precisely. Pick out the active points along this trajectory to construct the APMJ function (\ref{eq:japmj}), then the following motion planning model can be obtained to solve the exact optimal trajectory:
\begin{equation}
  \begin{aligned}
      \min          \quad &   \beta g(\bm{x}_i,\bm{x}_f,t_f) + (1-\beta)\int_0^{t_f} L(\bm{x},\bm{u}) \mathrm{d}t \\  
      \mathrm{s.t.} \quad &    \bm{\dot{x}}=\bm{f}(\bm{x},\bm{u}) \\
      \quad &    \bm{x}(0)=\bm{x}_i, \quad \bm{x}(t_f)=\bm{x}_f \\
      \quad &    \bm{x}\in \bm{X}, \qquad \ \ \bm{u}\in \bm{U} \\
      \quad & 1+\bm{c}^\mathrm{T}\bm{p}\ge \delta \\
      \quad &    \bm{Qp} = \bm{b},\quad \bm{p}\ge\bm{0}
  \end{aligned}
  \label{eq:apmjframwork}
\end{equation}
where the notations $\bm{Q}$,$\bm{p},\bm{c}$ need to be distinguished from those in (\ref{eq:makktframwork}). In this model, $\bm{Q}$,$\bm{p},\bm{c}$ correspond to the active points whose cardinality is very low. Besides, the multipliers $\bm{\lambda}$, $\bm{v}$ and the complementary and equilibrium constraints disappear, so this model is much more easy to solve.

\subsection{The BOMP Algorithm}
\label{sec:algorithm}
Combining the above theories and techniques, a two-stage BOMP algorithm framework is formulated for robot trajectory generation and optimization mission. The algorithm details are shown in \textbf{Algorithm 1}, where steps 1-10 correspond to the initial stage, and step 11 is the final stage. In the initial stage, the MJ function is used to construct the motion planning problem (\ref{eq:makktframwork}), to generate a collision-free approximate optimal trajectory and to pick out the active-points along the trajectory. As the relationship between the $\epsilon$-MAKKT point and the exact KKT point, a decreasing sequence $\epsilon$ is adopted in the initial stage to solve a series of approximate optimal trajectories. The final stage figures out an optimal trajectory using the APMJ function based model (\ref{eq:apmjframwork}). The PSOC method solves both problems (\ref{eq:makktframwork}) and (\ref{eq:apmjframwork}), and the discrete nodes number in the two stages are 15 and 30, respectively. In order to limit the length of this paper, the PSOC method details are omitted, and readers can consult the reference paper \cite{ross2012review}. The IPOPT \cite{wachter2006implementation} algorithm of version 3.12.9 is used to solve the discrete large scale NLP. The IPOPT algorithm is designed with special options, such as convergence tolerance $1e^{-12}$, MA86 linear solver and the acceptable termination conditions are more strict than the desired ones to avoid algorithm early termination. The IPOPT algorithm options are listed in Table \ref{tab:ipoptoptions}. In the end, we point out that the little trick, step 9, used in the BOMP algorithm reduces the collision avoidance constraint numbers dramatically. That is, if the robot is very far away from an obstacle, then the collision avoidance constraint does not need to be considered at that moment. The little trick picks out the right critical collision avoidance constraints, and these constraints are called active constraints in this paper.
\begin{algorithm}[hb]
    \caption{The BOMP algorithm} 
    \hspace*{0.02in} {\bf Input:}
    $\delta$, an $\epsilon$ sequence $\{0.1,\,0.01,\,0.001,\,\cdots\}$, an initial trajectory guess $\bm{\tau}_0$, the discrete nodes number $K$ in the PSOC method
    and the active constraints threshold $J_a$. \\
    \hspace*{0.02in} {\bf Output:}
    the optimal trajectory $\bm{\tau}^*$.
    \begin{algorithmic}[1]
        \State Let $k=0$, $K=15$.  
        \Loop
            \State Using $\epsilon_k$ to construct the problem (\ref{eq:makktframwork}). 
            \State Solve problem (\ref{eq:makktframwork}) by the PSOC method with the initial guess $\bm{\tau}_k$ to get a new trajectory $\bm{\tau}_{k+1}$.
            \If{the trajectory $\bm{\tau}_{k+1}$ is collision free} 
                \State Let $\bm{\tau}=\bm{\tau}_{k+1}$ and break.
            \EndIf
            \State Let $k \leftarrow k+1$.
        \EndLoop
        \State Let $K=30$ and get the interpolated trajectory $\bm{\tau}$. 
        \State Pick out the active constraints along the trajectory $\bm{\tau}$ according to the rule $1+\bm{c}^\mathrm{T}\bm{p}\le J_a$.
        \State Pick out the active points for the active constraints.
        \State Construct problem (\ref{eq:apmjframwork}) using the active points and solve it by the PSOC method with the initial guess $\bm{\tau}$ to get the optimal trajectory $\bm{\tau}^*$.
    \end{algorithmic}
\end{algorithm}
\begin{table}[ht]
  \small
  \centering
  \vspace{-1.5ex}
  \caption{The IPOPT algorithm options.}
  \vspace{1.5ex}
  \begin{tabular}{lm{0.55\columnwidth}l}
    \hline
    \hline
    Parameter       & Description                                                                                    & Setting \\ \hline
    tol             & Desired convergence tolerance                                                                  & $1e^{-12}$    \\ \hline
    a\_tol          & Acceptable convergence tolerance                                                               & $1e^{-16}$    \\ \hline
    cv\_tol         & \multirow{2}{0.6\columnwidth}{Desired/Acceptable threshold for the constraint violation}       & \multirow{2}{*}{$1e^{-12}$}    \\ 
    acv\_tol        &                                                                                                &    \\ \hline
    c\_tol          & \multirow{2}{0.6\columnwidth}{Desired/Acceptable threshold for the complementarity conditions} & \multirow{2}{*}{$1e^{-4}$} \\
    ac\_tol         &                                                                                                & \\ \hline
    d\_tol          & \multirow{2}{0.6\columnwidth}{Desired/Acceptable threshold for the dual infeasibility}         & \multirow{2}{*}{1} \\
    ad\_tol         &                                                                                                & \\ \hline
    solver          & Linear solver used for step computation                                                        & MA86 \\
    \hline
    \hline
  \end{tabular}
  \label{tab:ipoptoptions}
\end{table}

\section{The BOMP Model Verifications}
\label{sec:applications}
In this section, the simulations of the BOMP algorithm in autonomous parking problem are shown. The computational and precision benefits of the BOMP model over the area criterion (AC) based model \cite{li2015unified} and the circle approximation method are demonstrated. And a real experiment on Turtlebot3 is conducted. C++ code is programmed in Linux system, and simulations are conducted on an Intel Core i7-7700K CPU with 8GB RAM that runs at 4.20GHz.
\subsection{Simulation in Autonomous Parking}
\label{sec:parking}
The vehicle kinematics model (\ref{eq:kinematics}) and the optimization performance index (\ref{eq:objective}) are used in this section. As the parking process is in low speed and to bound the trajectory curvature and its derivative \cite{li2015unified}, the following mechanical and physical constraints are considered.
\begin{equation}
    \vert \upsilon \vert\le\upsilon_{\mathrm{max}},\quad \vert \alpha \vert\le\alpha_{\mathrm{max}},\quad \vert \omega \vert\le\omega_{\mathrm{max}} 
  \label{eq:mechanicas}
\end{equation}
And the initial and the final state constraints are treated as:
\begin{equation}
  \left\{
    \begin{aligned}
      & x(0)=x_0,\ y(0)=y_0,\ \theta(0)=\theta_0 \\ 
      & \upsilon(0)=0,\ \upsilon(t_f)=0,\ \vert \theta(t_f) \vert\le\epsilon_\theta  \\
      & \vert (A_y(t_f)+B_y(t_f)-SW)/2\vert\le\epsilon_p \\ 
      & \vert (C_y(t_f)+D_y(t_f)-SW)/2\vert\le\epsilon_p \\
      & \vert (A_x(t_f)+D_x(t_f)-SL)/2\vert\le\epsilon_p \\ 
      & \vert (B_x(t_f)+C_x(t_f)-SL)/2\vert\le\epsilon_p
    \end{aligned}
  \right.
  \label{eq:terminal}
\end{equation}
where $(x_0,\,y_0,\,\theta_0)$ is the vehicle initial configuration, $(A_x,\, A_y)$ is the coordinate of vehicle corner point $\bm{A}$ described in $\{\bm{W}\}$, $(B_x,\, B_y)$, $(C_x,\, C_y)$ and $(D_x,\, D_y)$ have the similar meanings, $\epsilon_p\ge0$ and $\epsilon_\theta\ge0$ denote the position deviation and the orientation deviation of the vehicle central axis with respect to the parking spot central axis, respectively. Therefore, the constraints (\ref{eq:kinematics}), (\ref{eq:mechanicas}) and (\ref{eq:terminal}) form the upper-level constraints of the BOMP model. 
\begin{figure*}[ht] 
    \begin{minipage}{0.5\linewidth}
        \centering 
        \includegraphics{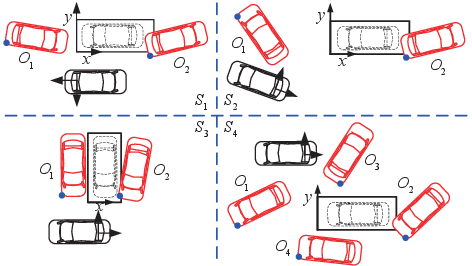}
        \caption{Four scenarios used to conduct the simulations. Red vehicles are obstacles, the black vehicle denotes the initial location, the dashed vehicles illustrate the ideal parking and each obstacle is the same size as the vehicle. The first two scenarios demonstrate the parallel parking and the third one demonstrates the vertical parking. While the last scenario is not usual in daily life, it's just designed to show the algorithm's capability.}
        \label{fig:scenarios}
    \end{minipage}
    \begin{minipage}{0.5\linewidth} 
        \centering 
        \captionof{table}{Transformation parameters $(x,\,y,\,\theta)$ of each obstacle with respect to the world coordinate system and the vehicle initial configuration $(x_0,\,y_0,\,\theta_0)$ in each case. The angle unit here is degree, but at other place it is radian.}
        \vspace{1.5ex}
        \small
        \renewcommand\arraystretch{0.5}
        \begin{tabular}{|r|c|l|c|l|}
            \hline
            \multirow{6}{*}{$S_1$} & \multirow{3}{*}{$O_1$} & \multirow{3}{*}{$(-5.6,\ 0.6,\ -10.0)$} & \multirow{2}{*}{$C_1$} & \multirow{2}{*}{$(-6.0,\ -2.75,\ 0.0)$} \\
                                   &                        &                                         &                        &                                         \\ \cline{4-5} 
                                   &                        &                                         & \multirow{2}{*}{$C_2$} & \multirow{2}{*}{$(0.0,\ -2.75,\ 0.0)$} \\ \cline{2-3}   
                                   & \multirow{3}{*}{$O_2$} & \multirow{3}{*}{$(5.8,\ -0.5,\ 13.0)$}  &                        &                                         \\ \cline{4-5}
                                   &                        &                                         & \multirow{2}{*}{$C_3$} & \multirow{2}{*}{$(0.0,\ -2.75,\ 180.0)$} \\  
                                   &                        &                                         &                        &                                         \\ \hline   

            \multirow{6}{*}{$S_2$} & \multirow{3}{*}{$O_1$} & \multirow{3}{*}{$(-7.0,\ 2.8,\ -55.0)$} & \multirow{2}{*}{$C_1$} & \multirow{2}{*}{$(-6.0,\ -2.0,\ 0.0)$} \\
                                   &                        &                                         &                        &                                         \\ \cline{4-5} 
                                   &                        &                                         & \multirow{2}{*}{$C_2$} & \multirow{2}{*}{$(-8.0,\ 0.0,\ -30.0)$} \\ \cline{2-3}   
                                   & \multirow{3}{*}{$O_2$} & \multirow{3}{*}{$(5.8,\ -0.5,\ 13.0)$}  &                        &                                         \\ \cline{4-5}
                                   &                        &                                         & \multirow{2}{*}{$C_3$} & \multirow{2}{*}{$(-4.0,\ -3.0,\ -36.0)$} \\  
                                   &                        &                                         &                        &                                         \\ \hline   

            \multirow{6}{*}{$S_3$} & \multirow{3}{*}{$O_1$} & \multirow{3}{*}{$(-2.2,\ 0.4,\ 0.0)$} & \multirow{2}{*}{$C_1$} & \multirow{2}{*}{$(-6.0,\ -2.0,\ 0.0)$} \\
                                   &                        &                                         &                        &                                         \\ \cline{4-5} 
                                   &                        &                                         & \multirow{2}{*}{$C_2$} & \multirow{2}{*}{$(1.0,\ -2.0,\ 0.0)$} \\ \cline{2-3}   
                                   & \multirow{3}{*}{$O_2$} & \multirow{3}{*}{$(2.3,\ 0.3,\ -5.0)$}  &                        &                                         \\ \cline{4-5}
                                   &                        &                                         & \multirow{2}{*}{$C_3$} & \multirow{2}{*}{$(-3.0,\ -2.0,\ 180.0)$} \\  
                                   &                        &                                         &                        &                                         \\ \hline   

            \multirow{12}{*}{$S_4$}& \multirow{3}{*}{$O_1$} & \multirow{3}{*}{$(-6.3,\ 0.3,\ 20.0)$} & \multirow{4}{*}{$C_1$} & \multirow{4}{*}{$(-5.0,\ 7.5,\ 0.0)$} \\
                                   &                        &                                       &                        &                                       \\ 
                                   &                        &                                       &                        &                                       \\ \cline{2-3} 
                                   & \multirow{3}{*}{$O_2$} & \multirow{3}{*}{$(6.8,\ -1.0,\ 40.0)$} &                        &                                          \\ \cline{4-5}
                                   &                        &                                       & \multirow{4}{*}{$C_2$} & \multirow{4}{*}{$(-5.0,\ 5.0,\ -10.0)$} \\ 
                                   &                        &                                       &                       &                                           \\ \cline{2-3}
                                   & \multirow{3}{*}{$O_3$} & \multirow{3}{*}{$(2.0,\ 3.6,\ 60.0)$} &                        &                                          \\ 
                                   &                        &                                       &                       &                                           \\ \cline{4-5}
                                   &                        &                                       & \multirow{4}{*}{$C_3$} & \multirow{4}{*}{$(-1.0,\ 6.0,\ 0.0)$} \\ \cline{2-3} 
                                   & \multirow{3}{*}{$O_4$} & \multirow{3}{*}{$(-1.5,\ -2.2,\ -3.0)$} &                       &                                           \\
                                   &                        &                                       &                       &                                           \\
                                   &                        &                                       &                       &                                           \\ \hline
        \end{tabular}
        \label{tab:transformation parameters}
    \end{minipage} 
\end{figure*}

In this application, four scenarios and three cases in each scenario are designed to reflect the generality, the robustness and the advantages of the BOMP algorithm. Fig. \ref{fig:scenarios} illustrates the four scenarios and one case in each scenario. Red vehicles are obstacles, the black vehicle denotes its initial location, and the dashed vehicle denotes the ideal parking. All the four scenarios correspond to irregularly placed obstacles that can be encountered easily in daily life and the differences between cases in each scenario are only the initial configurations of the vehicle (see Table \ref{tab:transformation parameters}, the abbreviations $S,\,O,\,C$ represent the scenario, the obstacle and the case, respectively). For simplicity, each obstacle is the same size as the vehicle and collision avoidance is equivalent to the plane rectangle pair is collision free. The transformation parameters $(x,\,y,\,\theta)$ of the obstacle coordinate system (it is originated at the blue corner point of the obstacle) relative to the world coordinate system are used to represent each obstacle compactly. These transformation parameters $(x,\,y,\,\theta)$ are listed in Table \ref{tab:transformation parameters}. 

\begin{figure*}[t] 
    \vspace{-3ex}
    \begin{minipage}{0.5\linewidth}
        \begin{figure}[H]
          \centering
          \includegraphics{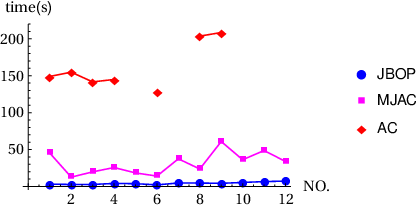}
          \caption{Total computation time of the BOMP, the MJAC and the AC algorithms used to solve the whole 12 simulation cases. Missing values in the AC algorithm mean that the algorithm fails to solve these problems.}
          \label{fig:time}
        \end{figure}
    \end{minipage}
    \begin{minipage}{0.5\linewidth}
        \begin{algorithm}[H]
            \caption{The AC algorithm} 
            \hspace*{0.02in} {\bf Input:}
            An initial trajectory guess $\bm{\tau}_0$, the discrete nodes number $K$ in the PSOC method. \\
            \hspace*{0.02in} {\bf Output:}
            The optimal trajectory $\bm{\tau}^*$.
            \begin{algorithmic}[1]
                \State Let $K=15$, using the area criterion constraint (\ref{eq:areacriterion}) to construct the motion planning problem and solve it by the PSOC method with the initial guess $\bm{\tau}_0$ to get a new trajectory $\bm{\tau}$. 
                \State Let $K=30$ and get the interpolated trajectory $\bm{\tau}$. 
                \State Construct the new problem and solve it by the PSOC method with the initial guess $\bm{\tau}$ to get the optimal trajectory $\bm{\tau}^*$.
            \end{algorithmic}
        \end{algorithm}

    \end{minipage}
\end{figure*}

\begin{figure*}[!ht]
  \vspace{0.5ex}
  \centering
  \subfigure[Vehicle motion trajectory]{\includegraphics{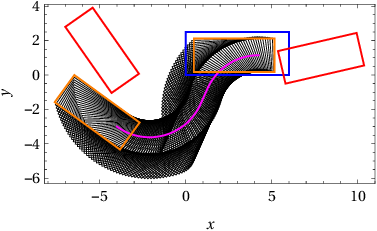}}
  \subfigure[State trajectories]{\includegraphics{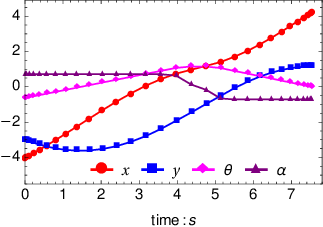}}
  \subfigure[Control trajectories]{\includegraphics{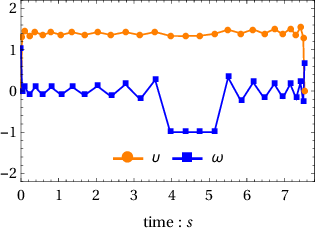}}
  \vspace{-2ex}
  \caption{Optimization results for scenario 2 case 3 by the BOMP algorithm. In (a), the orange rectangles represent the vehicle initial and final locations, and the magenta curve shows the motion trajectory of vehicle rear wheel axis mid-point. The state trajectories in (b) are very smooth, while the control trajectories in (c) are oscillating. The state and control trajectories in all other problems have the same characteristics.}
  \label{fig:scenario23}
\end{figure*}

\begin{figure*}[!ht]
  \vspace{-1ex}
  \centering
  \subfigure[Vehicle motion trajectory]{\includegraphics{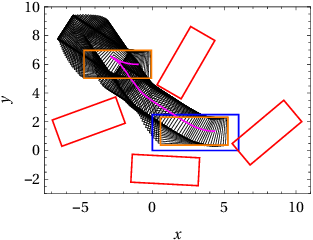}}
  \subfigure[State trajectories]{\includegraphics{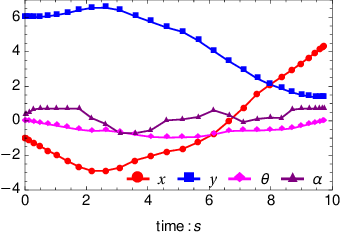}}
  \subfigure[Control trajectories]{\includegraphics{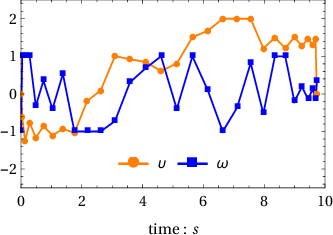}} 
  \vspace{-2ex}
  \caption{Optimization results for scenario 4 case 3 by the BOMP algorithm.}
  \label{fig:scenario43}
  \vspace{-2ex}
\end{figure*}

\begin{figure*}[!ht]
  \vspace{-1ex}
  \centering
  \subfigure[$S_2C_3$]{\includegraphics[width=0.38\textwidth]{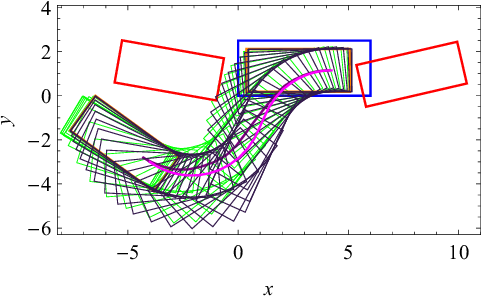}} 
  \subfigure[$S_3C_2$]{\includegraphics[width=0.2\textwidth]{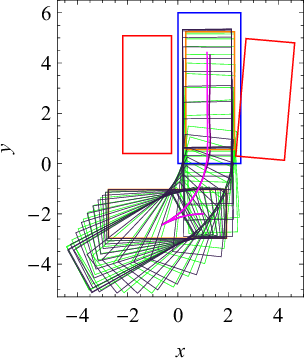}}
  \subfigure[$S_4C_1$]{\includegraphics[width=0.33\textwidth]{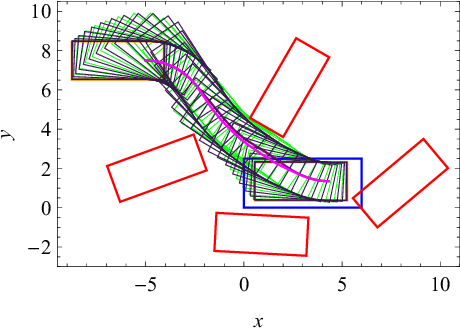}}
  \caption{Comparisons of the trajectory results between the two stages in the BOMP algorithm. The green and purple trajectories are the results from the initial stage, while the black and magenta trajectories are from the final stage. The results demonstrate that the final exact optimal trajectory is a small variation of the previous approximate optimal trajectory, and the APMJ function convergence strategy is appropriate.}
  \label{fig:stages}
\end{figure*}

The specified simulation parameters for the parking spot, the vehicle and the \textbf{Algorithm 1}'s input are listed in Table \ref{tab:simulation parameters}, where the vehicle size parameters, mechanical and physical constraints parameters are originated from \cite{li2015unified}. As for the initial trajectory guess $\bm{\tau}_0$ in \textbf{Algorithm 1}, it is specified very easily. Guess the parking completion time arbitrarily, then $\bm{\tau}_0$ is composed of $x(t)=x_0, y(t)=y_0, \theta(t)=\theta_0,\alpha(t)=0,\upsilon(t)=0,\omega(t)=0,\bm{p}(t)=\bm{0},\bm{\lambda}(t)=\bm{0}, \bm{v}(t)=\bm{0}$ for any $t$. At last, we point out that the coordinate system used for the $J_2$-function is established at the center of the vehicle, i.e., the inverse coordinate transformation is applied to transform the obstacles into vehicle center coordinate system. Under these conditions, the BOMP algorithm solves all these $4\times3$ problems with the same sequence $\epsilon\in\{0.1,\,0.01\}$. That is, through two iterations in the initial stage and one iteration in the final stage, the BOMP algorithm solves the whole autonomous parking problems. The values in the $\epsilon$ sequence are much larger than the algorithm convergence tolerance $1e^{-12}$, so the two problems in the initial stage are very easy to solve. The computation results are shown in Table \ref{tab:timetable} and Figs. \ref{fig:time}-\ref{fig:stages}.
\begin{table}[ht]
  \small
  \centering
  \vspace{-2ex}
  \caption{The specified parameters for the parking spot, the vehicle and \textbf{Algorithm 1}'s input.}
  \vspace{1.5ex}
  \begin{tabular}{lm{0.5\columnwidth}l}
    \hline
    \hline
    Parameter                 & Description                                                     & Setting \\ \hline
    SL                        & Parking spot length                                             & 6.00 m   \\ \hline
    SW                        & Parking spot width                                              & 2.50 m \\ \hline
    $l$                       & Wheel base length                                               & 2.800 m \\ \hline
    $l_1$                     & Front overhang length                                           & 0.960 m\\ \hline
    $l_2$                     & Rear overhang length                                            & 0.929 m\\ \hline
    $W$                       & Vehicle width                                                   & 1.942 m\\ \hline
    $\upsilon_{\mathrm{max}}$ & Bound of velocity                                               & 2.00 m/s \\ \hline
    $\alpha_{\mathrm{max}}$   & Bound of steering angle                                         & 0.714 rad \\ \hline
    $\omega_{\mathrm{max}}$   & Bound of angular velocity                                       & 1.00 rad/s    \\ \hline
    $\epsilon_{\mathrm{p}}$              & Bound of terminal position error                     & 0.10 m  \\ \hline
    $\epsilon_{\mathrm{\theta}}$         & Bound of terminal angle error                        & 0.17 rad \\ \hline
    $\delta$                  & The safety pseudodistance of the $J_2$-function                 & 0.05 \\ \hline
    $J_a$                     & The active constraints threshold                                & 0.30 \\

    \hline
    \hline
  \end{tabular}
  \label{tab:simulation parameters}
\end{table}

Then the computation complexity comparison between the BOMP algorithm and the AC algorithm is made. Given a point $\bm{b}$ and four corner points $\bm{A},\,\bm{B},\,\bm{C},\,\bm{D}$ of a rectangle in the plane, the AC collision avoidance constraint used in this paper is:
\begin{equation}
    \frac{S_{\triangle bAB}+S_{\triangle bBC}+S_{\triangle bCD}+S_{\triangle bDA}}{S_{\square ABCD}} \ge 1.025
    \label{eq:areacriterion}
\end{equation}
where $S_\triangle$ and $S_\square$ denote the area of a triangle and a rectangle, respectively. And there are eight nonlinear constraints (\ref{eq:areacriterion}) for a pair of plane rectangles collision avoidance problem. Then replacing (\ref{eq:framwork5}) and (\ref{eq:framwork6}) by these constraints, the AC based motion planning problem can be obtained, and the AC algorithm is shown in \textbf{Algorithm 2}. Correspondingly, steps 1-2 are called the initial stage, and step 3 is the final stage. Because the AC does not define distance explicitly, the active constraints filtrating process does not appear in \textbf{Algorithm 2}. Since in \cite{li2015unified}, the IPOPT options and the computation time were not shown, the algorithm options in Table \ref{tab:ipoptoptions} are used. And the initial guess $\bm{\tau}_0$ is selected the same way $x(t)=x_0,\, y(t)=y_0,\, \theta(t)=\theta_0,\,\alpha(t)=0,\,\upsilon(t)=0,\,\omega(t)=0$. However, the AC algorithm does not solve the whole autonomous parking problems and it fails in the initial stage. The results are shown in Table \ref{tab:timetable} and Fig. \ref{fig:time}.
\begin{table*}[t]
    \footnotesize
    \centering
    \caption{The computation results comparison between the BOMP, the MJAC and the AC algorithms, where I and F denote the computation time in the initial stage and final stage, respectively, T is the total computation time and $v^*$ is the optimal value. The MJAC algorithm has the same computation time in the initial stage as the BOMP algorithm. The symbol $\times$ denotes that the AC algorithm fails to solve the corresponding problem, or more precisely, it all fails in the initial stage.}
    \vspace{1.5ex}
    \begin{tabular}{|c|c|c|c|c|c|c|c|c|c|c|c|c|c|}
        \hline
                        \multicolumn{2}{|c|}{}  & \multicolumn{4}{c|}{BOMP}         &  \multicolumn{4}{c|}{MJAC} &  \multicolumn{4}{c|}{AC} \\
    \cline{3-14}    \multicolumn{2}{|c|}{}  &  I (s)   &  F (s)   &  T (s)   & $v^*$        &   I (s)     &   F (s)   &   T (s)   & $v^*$    &    I (s)     &    F (s)     &    T (s)     & $v^*$ \\ \hline 
        \multirow{4}{*}{$S_1$}  & $C_1$         & 2.82 & 0.30 & 3.12 & 17.14        & $\ast$  & 42.46 & 45.28 & 16.52    &  86.43   &  62.61   &  149.04  & 16.98  \\
        \cline{2-14}            & $C_2$         & 2.45 & 0.27 & 2.72 & 13.07        & $\ast$  & 10.20 & 12.65 & 12.76    &  96.34   &  58.50   &  154.84  & 14.64 \\
        \cline{2-14}            & $C_3$         & 2.41 & 0.36 & 2.77 & 17.33        & $\ast$  & 17.40 & 19.81 & 17.18    &  83.15   &  58.47   &  141.62  & 17.58 \\ \hline
        \multirow{3}{*}{$S_2$}  & $C_1$         & 3.85 & 0.30 & 4.15 & 15.73        & $\ast$  & 21.86 & 25.71 & 15.41    &  94.01   &  50.95   &  144.96  & 15.26 \\
        \cline{2-14}            & $C_2$         & 3.55 & 0.27 & 3.82 & 19.40        & $\ast$  & 14.81 & 18.36 & 19.13    & $\times$ & $\times$ & $\times$ & $\times$ \\
        \cline{2-14}            & $C_3$         & 1.56 & 0.62 & 2.18 & 14.88        & $\ast$  & 12.54 & 14.10 & 14.88    & 80.82 & 46.84 & 127.66 & 14.88 \\ \hline
        \multirow{3}{*}{$S_3$}  & $C_1$         & 4.25 & 0.49 & 4.74 & 17.36        & $\ast$  & 33.22 & 37.47 & 16.65    & $\times$ & $\times$ & $\times$ & $\times$ \\
        \cline{2-14}            & $C_2$         & 4.36 & 0.32 & 4.68 & 14.55        & $\ast$  & 19.34 & 23.70 & 13.73    & 132.64 & 71.00 & 203.64 & 13.57 \\
        \cline{2-14}            & $C_3$         & 3.82 & 0.44 & 4.26 & 19.65        & $\ast$  & 57.08 & 60.90 & 18.11    & 123.78 & 84.73 & 208.51 & 18.46 \\ \hline
        \multirow{3}{*}{$S_4$}  & $C_1$         & 5.05 & 0.45 & 5.50 & 16.58        & $\ast$  & 31.18 & 36.23 & 16.39    & $\times$ & $\times$ & $\times$ & $\times$ \\
        \cline{2-14}            & $C_2$         & 6.03 & 0.59 & 6.62 & 17.81        & $\ast$  & 42.68 & 48.11 & 14.64    & $\times$ & $\times$ & $\times$ & $\times$ \\
        \cline{2-14}            & $C_3$         & 6.95 & 0.35 & 7.30 & 18.35        & $\ast$  & 26.92 & 33.87 & 15.89    & $\times$ & $\times$ & $\times$ & $\times$ \\ \hline
    \end{tabular}
    \label{tab:timetable}
\end{table*}

According to the concept that a near-optimal initial guess always facilitates the solving of a complicated problem. And in order to show the impact of the active constraints filtrating process, the collision avoidance constraints in the BOMP algorithm's final stage (step 11) are just replaced by the AC based method. That is, the MJ function in the initial stage generates a collision-free approximate optimal trajectory $\tau$, picks out the active constraints, and then the AC method figures out the optimal trajectory using $\tau$ as the initial guess. This algorithm is called the MJAC algorithm. Under this circumstance, the MJAC algorithm solves all these $4\times3$ problems, and the results are shown in Table \ref{tab:timetable} and Fig. \ref{fig:time}.

The success rates of the MJAC algorithm and the AC algorithm show that a good initial guess reduces the difficulty of the AC algorithm. The computation time in the final stage of the MJAC algorithm and the AC algorithm demonstrates that the active-constraints filtrating process accelerates the computation speed several times. From Table \ref{tab:timetable}, we can see that the shortest and the longest time of the BOMP algorithm to find an optimal collision-free trajectory for the 12 problems are 2.18 and 7.30 seconds, respectively. For the MJAC algorithm, they are 12.65 and 60.90 seconds. While for the AC algorithm in the solved problems, they are 127.66 and 208.51 seconds. The reasons why the BOMP algorithm has significant superiorities in computation speed may be: 
\begin{itemize}[itemindent=-0.5mm,leftmargin=4mm,topsep=1pt,itemsep=0pt,parsep=1pt]
  \item $J_2$-function collision avoidance is linear programming, while the AC based collision avoidance method needs to calculate the distance between points, then the area of a triangle, such that it has a high degree of non-linearity. Linear programming is the most well-studied problem, and finding the solution of it is intrinsically much faster than the problem with nonlinear constraints.
  \item Variables in the $J_2$-function make the BOMP algorithm only concern the closest components implicitly in the initial stage. However, the AC based collision avoidance method needs to consider the eight points between a pair of rectangles simultaneously and explicitly, whose optimization direction is difficult to determine.
  \item The BOMP algorithm finds a sequence of solution progressively approaching the optimal solution such that it significantly reduces the difficulty to solve the problem, while the AC method uses a global step to find the solution which tends to break down in complicated problems. 
\end{itemize}

Limited by the length of this paper, the optimized vehicle motion and the optimized trajectories of state and control variables by the BOMP algorithm are just partly plotted in Figs. \ref{fig:scenario23}-\ref{fig:stages}. The results demonstrate that the BOMP algorithm is capable of handling parking motion planning problems successfully and efficiently. Besides, the BOMP algorithm can autonomously park a vehicle in a much more complicated scenario that needs several maneuvers like a skillful driver. Meanwhile, the terminal orientation deviation and the position deviation are considered, so the vehicle stops as parallel and close as to parking spot central axis. It is worth emphasizing that all these $4\times3$ simulations are conducted without adjustment of any algorithm options, like $\epsilon$, and this shows the robustness, generality, and unification of the BOMP model and the BOMP algorithm. However, because the vehicle kinematics, mechanical and physical constraints in our application is similar to that of \cite{li2015unified}, the control variables smoothness issue and terminal steering angle issue also exist in the BOMP algorithm, readers may consult that paper for the reasons. These issues are inherent in the modeling process and need some unique but not complicated techniques to overcome. Nevertheless, this paper focuses on the BOMP model and the solution to it, so these issues are left to resolve.

\subsection{Polygon VS Circle Approximation}
\label{sec:overlapping}
\begin{figure}[t]
    \centering
    \subfigure[]{\includegraphics{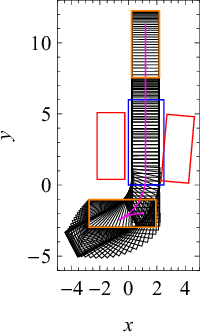}}
    \subfigure[]{\includegraphics{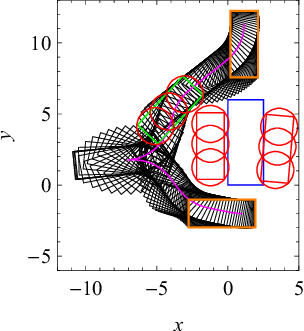}}
    \caption{The vehicle motion obtained by the BOMP algorithm and the circle approximation method, respectively. The green rectangle in (b) represents the vehicle at an instant moment, and the three overlapping circles represent the approximation of the vehicle and the obstacles.}
    \vspace{-2ex}
  \label{fig:Jcvertical}
\end{figure}

One motion planning problem in scenario 3 of the previous section is used to illustrate the different impacts of polygon and circle approximation methods. The planning task is to let the vehicle move from the initial configuration $(1.0,\,-2.0,\,0.0)$ to the final configuration $(1.0,\,10.0,\,\pi/2.0)$. Fig. \ref{fig:Jcvertical}(a) shows the result of the BOMP algorithm, and the vehicle and obstacles are still represented as rectangles. Fig. \ref{fig:Jcvertical}(b) shows the result of circle approximation method. The vehicle and the obstacles are approximated by three overlapping circles of radius 1.3m (the red circles), respectively. The results demonstrate that the circle approximation makes the object expanding, and the vehicle can not pass through narrow passages. To approximate the object exactly, there should be many more different radius circles, and this will increase the computation time dramatically. The drawbacks of circle approximation will become increasingly severe if the ratio of rectangle width and length reduces further. Because of the high approximation precision and low overlapping rate of the polyhedron or polygon approximation method, the simplicity of $J_2$-function and the fast computation speed of the APMJ function, the BOMP model will compensate motion planning methods in many domains. 

\vspace{-1ex}
\subsection{Physical Experiment}
\label{sec:physical}
\begin{figure}[t]
  \centering
  \includegraphics[width=0.35\textwidth]{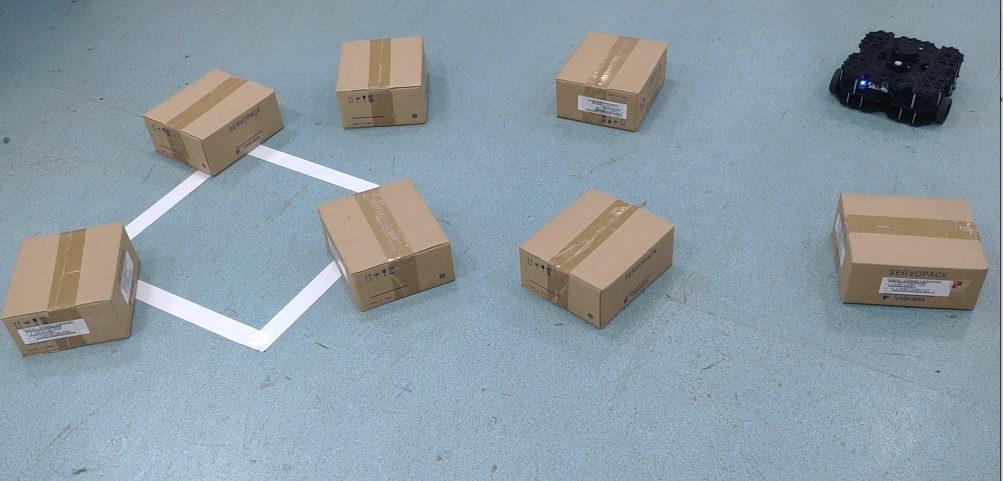}
    \caption{The environment setup for this experiment. There are seven obstacles in a very small area ($1.5m\times1.5m$) and the goal of all the experiments is to let Turtlebot3 stop at the center of the white square which is $0.5m\times0.5m$.}
  \label{fig:tbexperiment}
\end{figure}

\begin{figure}[!t]
  \centering
  \includegraphics{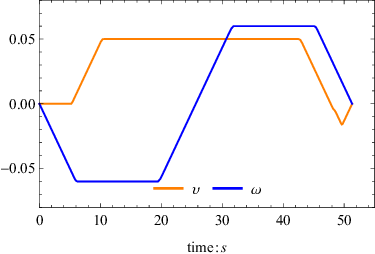}
  \caption{Given robot initial location, the calculated global control variables trajectories by the BOMP algorithm.}
  \label{fig:tbexperiment_controls}
\end{figure}

We performed a physical experiment involving a mobile robot (TurtleBot3) to verify the correctness and effectiveness of the BOMP model (see in Fig. \ref{fig:tbexperiment}). To simplify the experiment, the robot dynamics was not considered, and the robot moved at very low speed. Using SLAM technique, the environment geometry can be obtained and then through convex decomposition \cite{mamou2009simple} the polygons information are available. Then given an initial robot location, the global control commands (linear and angular speed) are calculated by the BOMP algorithm. In this experiment, the time-optimal trajectory is sought, and the global control variables trajectories corresponding to Fig. \ref{fig:tbexperiment} are shown in Fig. \ref{fig:tbexperiment_controls}. When the robot is executing the control commands, the dynamic uncertainty and drift will accumulate the driving error. When this error exceeds tolerance, replanning process is triggered. The details of robot odometry, localization, control framework, error accumulation calculation, and replanning cycles will be presented in our future works. The videos of this experiment are available at http://www.hust.edu.cn.

\section{Conclusion and Prospects}
\label{sec:conclusion}
In this paper, we have proposed a general and unified BOMP model for robot trajectory generation and optimization in obstacles environment, which makes the optimal control method a practical approach for complex high-precision robot motion planning problems. The upper-level optimal control is designed for robot nonlinear dynamics, while the lower-level $J_2$-function is for geometry collision-free constraint. Simulations in complex autonomous parking scenarios and experiment on Turtlebot3 demonstrate the computation superiorities and efficiency of the BOMP model. The highlights of this paper lie in the following aspects.
\begin{enumerate}[leftmargin=4mm,topsep=1pt,itemsep=0pt,parsep=1pt]
  \item The BOMP model utilizes the $J_2$-function linear programming to avoid collision. Because of the simplicity of the $J_2$-function, the BOMP model can solve robot motion planning problems in very high speed and high precision. 
  \item The BOMP model is an open framework that any user-specified constraints can be incorporated into the upper-level or the lower-level constraints. And the BOMP model is dimensionality independent. Therefore, this model can be utilized in manipulators, vehicles and humanoid robots, etc. 
  \item The BOMP algorithm makes full use of the convergence property of the MAKKT theory and applies the MJ function and the APMJ function in separate stages. This convergence strategy makes the BOMP algorithm very fast. 
\end{enumerate}

However, many aspects need to be studied deeply and comprehensively. First, one main issue is the computation efficiency for robots moving in cluttered and obstacles intensive environment. The future studies should be concentrated on the initial stage of the algorithm, and one possible solution is to combine the previous research theory, like sampling based methods. While in real applications, some simple techniques can be used. Take the parking problem as an example \cite{li2016time}, make finite classifications of the parking cases and calculate the trajectory offline to construct a database. Then in real parking process, pick out a closest recorded solution to the current situation to initialize the algorithm and finally solve the problem online. Second, motion planning for robots in dynamical environment with moving obstacles are regarded as the most difficult, important and significant research field. Whereas the BOMP model only solves the static problem at present, there are still extensive works to adopt the obstacles movement prediction and estimation theory into the BOMP framwork. Third, with appropriate objectives and constraints in different levels, the multilevel problems have a close connection to multiple objectives optimization problems. However, solving the multilevel problem is much more difficult than the bilevel problem, and the related theories have not been proposed. All these three unresolved issues are challenging and meaningful for the development of intelligent robots and will be discussed in our future works. Moreover, despite the simulation results and simple experiment on Turtlebot3 in this paper, extensive work is needed to conduct experiments on real complex applications and to verify the proposed algorithm.


\ifCLASSOPTIONcaptionsoff
  \newpage
\fi

\bibliographystyle{IEEEtran}
\bibliography{IEEEabrv,article}

\vspace{-2ex}
\appendix
\renewcommand{\appendixname}{Appendix~\Alph{section}}
\section{Appendix}
In this appendix, we demonstrate the complexity of the BOMP model scales with a growing number of obstacles. Without loss of generality, we make assumptions that $\bm{x}\in\Re^{n_1}$, $\bm{u}\in\Re^{n_2}$, $\bm{p}\in\Re^{n_3}$ and $\bm{Q}\in\Re^{m\times n_3}$ in (\ref{eq:framwork}). Then in the converted single level optimal control model (\ref{eq:makktframwork}), $\lambda\in\Re^{n_3}$ and $v\in\Re^m$. 
In PSOC method, the discrete nodes are selected as Legendre-Gauss collocation points or Chebyshev-Gauss collocation points which are on interval $[-1,\,1]$. Take the Legendre-Gauss collocation points as example in this paper, given the $K+1$ discrete nodes $[\tau_0,\tau_1,\cdots,\tau_K]$ in interval $[-1,\,1]$, then discretize $\bm{x}$, $\bm{u}$, $\bm{p}$, $\bm{Q}$ and $\bm{f}(\bm{x},\bm{u})$ as:
\begin{equation}
    \begin{aligned}
    & \bm{x}_k=\bm{x}(\frac{t_f}{2}+\tau_k\frac{t_f}{2}), \quad \bm{u}_k=\bm{u}(\frac{t_f}{2}+\tau_k\frac{t_f}{2}) \\
    & \bm{p}_k=\bm{p}(\frac{t_f}{2}+\tau_k\frac{t_f}{2}), \quad \bm{Q}_k=\bm{Q}(x_k), \\
    & \bm{f}_k=\bm{f}(\bm{x}_k,\,\bm{u}_k) \\ 
    & k=0,1,2,\cdots,K 
    \end{aligned}
    \label{eq:discrete}
\end{equation}
And let $\bm{X}=[\bm{x}_0,\,\bm{x}_1,\,\cdots,\,\bm{x}_K]$, $\bm{F}=[\bm{f}_0,\,\bm{f}_1,\,\cdots,\,\bm{f}_K]$, then $\bm{X}\in\Re^{n_1\times(K+1)}$, $\bm{F}\in\Re^{n_1\times(K+1)}$. Restrict the constraints to be satisfied at each discrete time node, a large scale sparse nonlinear optimization problem of (\ref{eq:makktframwork}) is obtained: 
\begin{subequations}
  \begin{align}
      \min   \quad &      \beta g(\bm{x}_i,\,\bm{x}_f,t_f) + (1-\beta) \frac{t_f}{2}\sum_{k=0}^{K} \omega_k L(\bm{x}_k,\,\bm{u}_k) \nonumber \tag{26} \\ 
      \label{eq:discretkktcontinuousegeneralbilevel1} \mathrm{s.t.} \quad &     \bm{X}\bm{D}=\frac{t_f}{2}\bm{F} \\
      \label{eq:discretkktcontinuousegeneralbilevel2}               \quad &     \bm{x}_0=\bm{x}_i, \quad \bm{x}_K=\bm{x}_f \\
      \label{eq:discretkktcontinuousegeneralbilevel3}               \quad &     \bm{x}_k\in \bm{X}, \qquad \ \ \bm{u}_k\in \bm{U} \\
      \label{eq:discretkktcontinuousegeneralbilevel4}               \quad &     1+\bm{c}^\mathrm{T}\bm{p}_k\ge\epsilon+\delta \\
      \label{eq:discretkktcontinuousegeneralbilevel5}               \quad &     \bm{Q}_k\bm{p}_k=\bm{b},\quad \bm{p}_k\ge\bm{0},\quad \bm{\lambda}_k\ge\bm{0} \\
      \label{eq:discretkktcontinuousegeneralbilevel6}               \quad &     \Vert\bm{c}-\bm{\lambda}_k+\bm{Q}_k^\mathrm{T}\bm{v}_k\Vert\le\sqrt{\epsilon},\quad \bm{\lambda}_k^\mathrm{T}\bm{p}_k\le\epsilon \\
                                                                    \quad & k=0,1,2,\cdots,K \nonumber 
  \end{align}
  \label{eq:vectordiscretkktcontinuousegeneralbilevel}
\end{subequations}
\!\!where $\bm{D}\in\Re^{(K+1)\times(K+1)}$ is a constant matrix and $\omega_k$ is a constant scalar.

Therefore, let $K^\prime=K+1$, there are $n_1*K^\prime$ variables $\bm{x}_k$, $n_2*K^\prime$ variables $\bm{u}_k$, $n_3*K^\prime$ variables $\bm{p}_k$, $n_3*K^\prime$ variables $\bm{\lambda}_k$, $m*K^\prime$ variables $\bm{v}_k$ and one variable $t_f$.
The constraint numbers in respect to the constraints (\ref{eq:discretkktcontinuousegeneralbilevel1})-(\ref{eq:discretkktcontinuousegeneralbilevel6}) are shown in Table \ref{tab:constraintnumbers}. The total optimization variables number and constraints number are $(n_1+n_2+2n_3+m)*K^\prime+1$ and $(2n_1+n_2+2n_3+m+3)*K^\prime+2n_1+1$, respectively, where the number one comes from $t_f>0$. They are always very large, so the converted nonlinear optimization problem (\ref{eq:vectordiscretkktcontinuousegeneralbilevel}) is difficult to solve. In the autonomous parking application in Section \ref{sec:applications}, there are four state variables and two control variables. And in the MJ function for a plane rectangle pair collision avoidance problem, there are eight variables, three equality constraints and eight inequality constraints (the other inequality constraint is no active, so it is omitted). So in the autonomous parking application, consider a plane rectangle pair collision avoidance problem first, construct the motion planning and control problem (\ref{eq:makktframwork}), discretize the constraints (\ref{eq:makktframwork4})-(\ref{eq:makktframwork3}) to get a large scale nonlinear optimization problem, then $n_1=4,\, n_2=2, \, n_3=8, \, m=3$ and there are $6K^\prime+1+19K^\prime$ variables and $10K^\prime+9+22K^\prime$ constraints, respectively. If there are $n$ pairs of rectangle collision avoidance constraints, the variable and constraint numbers are $6K^\prime+1+19nK^\prime$ variables and $10K^\prime+9+22nK^\prime$ constraints, respectively. That is, the BOMP model is linear complexity with the obstacles number. In the scenario four application, there are four rectangle pairs and let $K^\prime=15$, then the variables number and constraints number are $6*15+19*15*4+1=1231$ and $10*15+9+22*15*4=1479$, respectively.
\begin{table}
    \centering
    \vspace{-2.0ex}
    \caption{The constraint numbers in problem (\ref{eq:vectordiscretkktcontinuousegeneralbilevel}), where $K^\prime=K+1$.}
    \vspace{1.0ex}
    \begin{tabular}{|c|c|c|c|}
        \hline
        Label & Numbers & Label & Numbers \\ \hline 
        (\ref{eq:discretkktcontinuousegeneralbilevel1}) & $n_1*K^\prime$   & (\ref{eq:discretkktcontinuousegeneralbilevel4}) &  $K^\prime$  \\ \hline 
        (\ref{eq:discretkktcontinuousegeneralbilevel2}) & $2n_1$        & (\ref{eq:discretkktcontinuousegeneralbilevel5}) & $(m+2n_3)*K^\prime$  \\ \hline 
        (\ref{eq:discretkktcontinuousegeneralbilevel3}) & $(n_1+n_2)*K^\prime$ & (\ref{eq:discretkktcontinuousegeneralbilevel6}) & $2K^\prime$ \\ \hline
    \end{tabular}
    \vspace{-2.0ex}
    \label{tab:constraintnumbers}
\end{table}

\vspace{-7ex}
\begin{IEEEbiography}[{\includegraphics[width=1in,height=1.25in,clip,keepaspectratio]{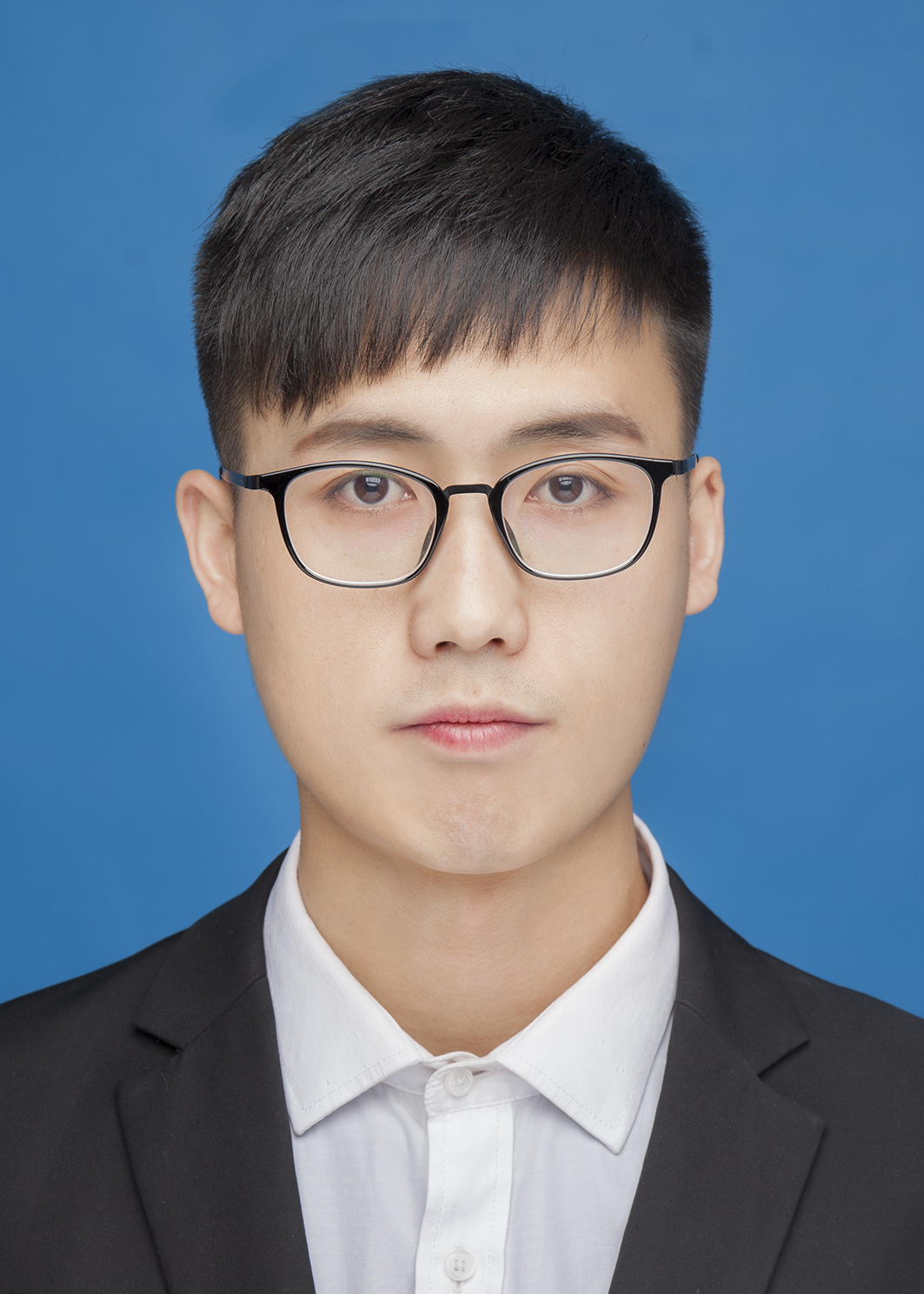}}]{Shenglei Shi} received  the B.S. degree in mechanical engineering from Huazhong University of Science and Technology (HUST), Wuhan, China, in 2016. He is currently  pursuing the Ph.D. degree at HUST.

His current research interests include motion planning and trajectory optimization in robotics.
\end{IEEEbiography}
\vspace{-7ex}
\begin{IEEEbiography}[{\includegraphics[width=1in,height=1.25in,clip,keepaspectratio]{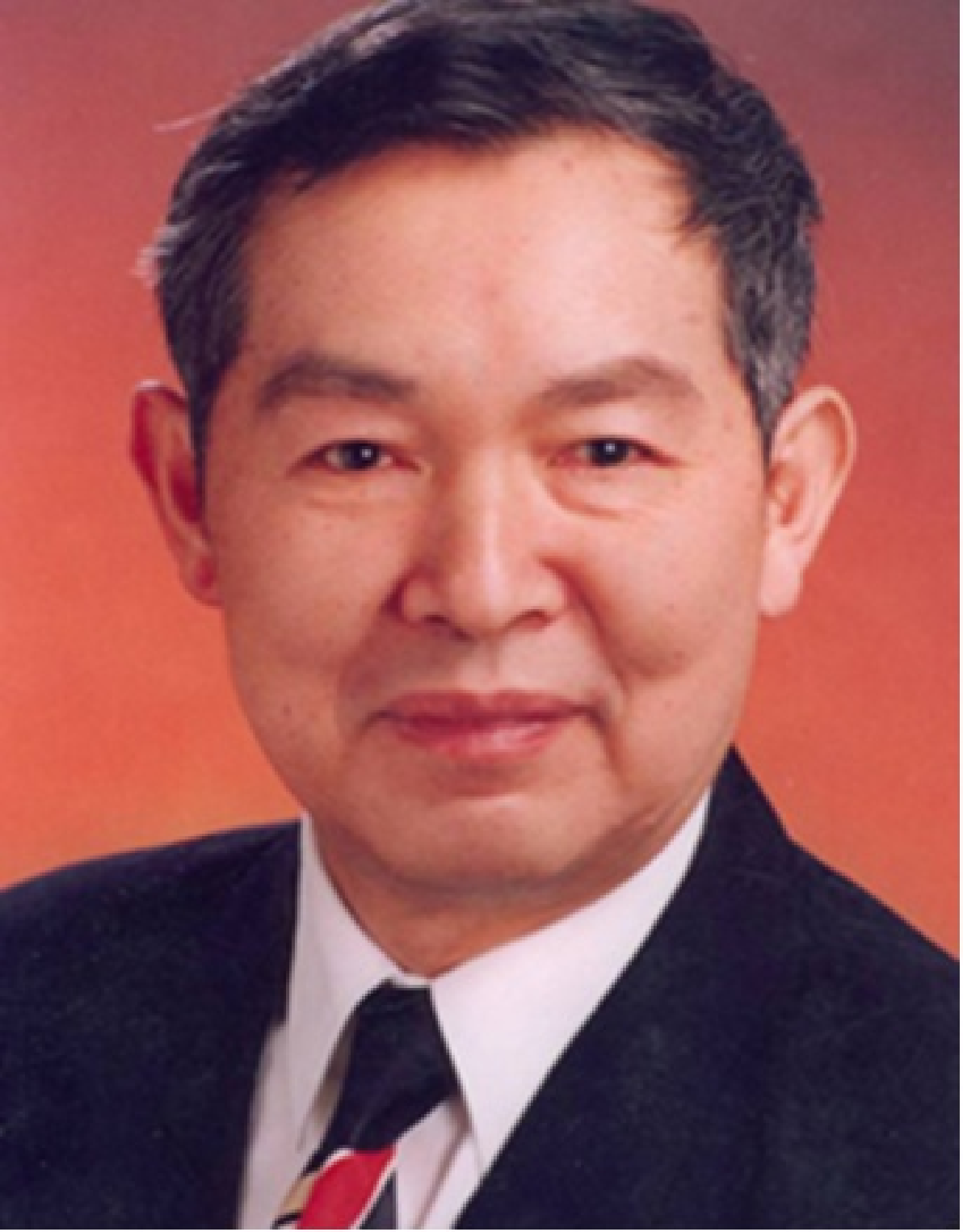}}]{Youlun Xiong} graduated from the Department of Mechanical Engineering and Postgraduate School at Xi¡¯an Jiatong University, Xi¡¯an, China, in 1962 and 1966, respectively.From 1966 to 1980 he was employed by the Department of Mechanical Engineering at Huazhong University of Science and Technology (HUST), Wuhan, China. From 1980 to 1982, he was a visiting scholar in the Department of Control Engineering at Sheffield University, Sheffield, UK. In 1982, he returned to the Department of Mechanical Engineering at HUST, where he is a Professor. From 1988 to 1989, he was a visiting professor in the Department of Aeronautical and Mechanical Engineering at the University of Salford, Salford, UK. In 1995 he was elected as Academician of Chinese Academy of Sciences.

His research interests include robotics, precision measurement, intelligence manufacturing and  manufacturing automation.
\end{IEEEbiography}
\vspace{-7ex}
\begin{IEEEbiography}[{\includegraphics[width=1in,height=1.25in,clip,keepaspectratio]{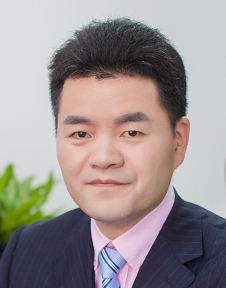}}]{Jiankui Chen} received the B.S. degree in vehicleoperation engineering from the Wuhan University of Technology, Wuhan, China, in 2001, and the M.S. and the Ph.D. degrees in mechatronic engi-neering from Huazhong University of Science and Technology (HUST), Wuhan, in 2006 and 2010, respectively. He  was a Post-Doctoral Fellow with the Department of Control Science and Engineering, HUST,in 2011. He is currently a Lecturer with the State Key Laboratory of Digital Manufacturing Equipmentand Technology, HUST.

His current research interests include flexible electronics manufacturing and roll-to-roll Web transport system control.
\end{IEEEbiography}
\vspace{-7ex}
\begin{IEEEbiography}[{\includegraphics[width=1in,height=1.25in,clip,keepaspectratio]{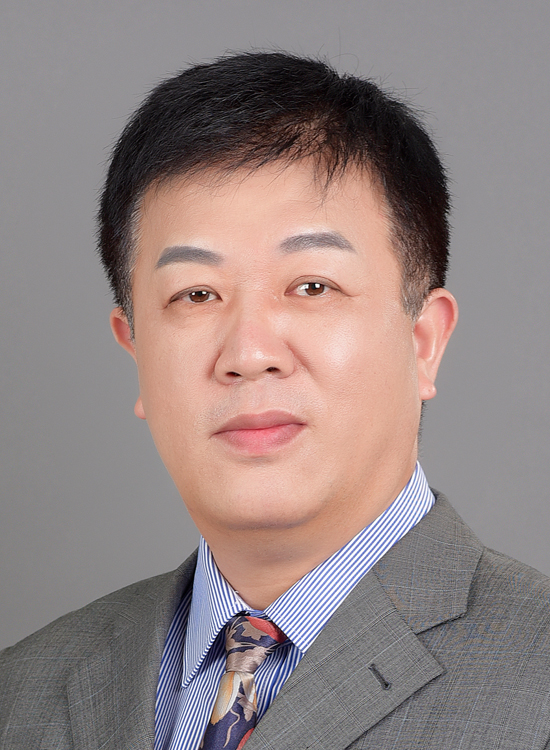}}]{Caihua Xiong} received the Ph.D. degree in mechanical engineering from Huazhong University of Science and Technology (HUST), Wuhan, China, in 1998. From 1999 to 2003, he was with City University of Hong Kong and The Chinese University of Hong Kong as a Postdoctoral Fellow, and with Worcester Polytechnic Institute, Worcester, MA, USA, as a Research Scientist. He is a Chang Jiang Professor with HUST. He has received the National Science Fund for Distinguished Young Scholars of China.

His research interests include biomechatronic prostheses, rehabilitation robotics, and robot motion planning and control.
\end{IEEEbiography}

\end{document}